\documentclass{naturep}
\usepackage[utf8]{inputenc}
\usepackage{lineno}
\usepackage{setspace}
\usepackage{color}
\usepackage{graphicx}
\usepackage{booktabs}
\usepackage{amssymb}
\usepackage{amsmath}
\usepackage{hyperref}
\usepackage{adjustbox}
\usepackage{subcaption}
\usepackage{tablefootnote}
\usepackage[margin=0.6in]{geometry}
\usepackage{verbatim}
\usepackage{caption}
\usepackage{paralist}
\usepackage{cleveref}
\usepackage{multirow}
\usepackage{tabularx} 
\usepackage{array} 
\usepackage{longtable}

\newcommand{\ours}{\textsc{STP-Bench}}
\newcommand{\oursint}{\textsc{STP-Bench-Internal}}
\newcommand{\oursext}{\textsc{STP-Bench-External}}
\newcommand{\nmodels}{21}
\newcommand{\nspotint}{609,024}
\newcommand{\nslideint}{202}
\newcommand{\nspotext}{185,831}
\newcommand{\nslideext}{73}

\newcommand\Heading[1]{
  \noindent\textbf{\Large{#1}}
}

\newcommand\heading[1]{
  \noindent\textbf{\large{#1}}
}

\title{\raggedright{\textbf{\ours: A Unified Systematic Benchmark for Virtual Spatial Transcriptomics from Histopathology Images}}}

\author{Youngmin Chung$^{1,2,\ast}$, Ji Hun Ha$^{3,\ast}$, Andrew H. Song$^{2,4,\ast}$, Cristina Almagro-Pérez$^{2,5,6,7}$, Chaeyoung Seo$^{1}$, Won Jun Suh$^{3}$, Jeong Won Beom$^{3}$, Kyoung Bin Oh$^{3}$, Eytan Ruppin$^{8,\dag}$, Faisal Mahmood$^{2,5,6,9,\dag}$, Joo Sang Lee$^{1,3,10\dag}$}
\date{}
    
\makeatletter
\let\saved@includegraphics\includegraphics
\AtBeginDocument{\let\includegraphics\saved@includegraphics}

\makeatother

\begin{document}
\maketitle
\begin{affiliations}
 \item Department of Artificial Intelligence, Sungkyunkwan University, Suwon, Republic of Korea
 \item Department of Pathology, Mass General Brigham, Harvard Medical School, Boston, MA, USA
 \item Department of Precision Medicine, School of Medicine, Sungkyunkwan University, Suwon, Republic of Korea
 \item Department of Translational Molecular Pathology and Division of Pathology and Laboratory Medicine, The University of Texas MD Anderson Cancer Center, Houston, TX, USA
 \item Data Science Program, Dana-Farber Cancer Institute, Boston, MA, USA
 \item Cancer Program, Broad Institute of Harvard and MIT, Cambridge, MA, USA
 \item Harvard-MIT Division of Health Sciences and Technology, Massachusetts Institute of Technology, Cambridge, MA, USA
 \item Translational Research Unit, Cedars-Sinai Medical Center, Los Angeles, CA, USA
 \item Harvard Data Science Initiative, Harvard University, Cambridge, MA, USA
 
\item Department of Digital Health, Samsung Advanced Institute of Health Science and Technology, Sungkyunkwan University, Seoul, Republic of Korea

 \item[$^{\ast}$] Contributed equally (Co-first)
 \item[$^{\dag}$] Co-senior authors \\
\textbf{Lead Contact}: \\
Joo Sang Lee (joosang.lee@skku.edu)
 \end{affiliations}

\clearpage
\Heading{Abstract}

\begin{spacing}{1.2}
\noindent
\textbf{Spatial transcriptomics (ST) provides unprecedented insights into tumor heterogeneity by capturing spatially resolved gene expression, yet its high experimental cost hinders large-scale adoption. Consequently, computational approaches that predict spatial gene expression directly from hematoxylin and eosin slides, termed virtual ST, have rapidly emerged. These approaches span diverse modeling paradigms, mirroring broader advances in predictive deep learning, and have demonstrated promising performance. Despite this progress, it remains difficult to properly assess the advances in the field, due to insufficient benchmarking: prior studies rely on small, heterogeneous datasets, inconsistent training and inference pipelines, and limited evaluation of biological interpretability and model robustness. To address these gaps, we present $\ours$, an extensive and standardized benchmark for comprehensive evaluation of virtual ST models. $\ours$ comprises six cancer types spanning two major ST platforms (Visium and Xenium), with each training dataset containing more than 30,000 spots and at least 15 slides to ensure statistical reliability. We evaluate \nmodels\ predictive approaches, all re-implemented with a unified pathology foundation model as the morphological encoder when architecturally applicable. Beyond conventional benchmarks that report average predictive accuracy on highly variable genes, we systematically examine which genes and gene sets are recoverable from histomorphology. We further evaluate the downstream biological utility of predicted profiles through cell-type deconvolution and spatial domain identification, and assess model reliability under domain shifts and data scaling. Notably, we find that unified morphological encoding substantially re-orders the model rankings established in prior studies, indicating that architectural innovations and image encoding have been conflated in previous evaluations. We publicly release $\ours$ to support reproducibility and to serve as a community benchmark at \url{https://github.com/NEXGEM/STP-Bench}.}

\end{spacing}


\clearpage
\begin{spacing}{1.35}
\Heading{Main}

Spatial transcriptomics (ST) has transformed cancer research by resolving gene expression profiles within its native tissue context \cite{staahl2016visualization,rao2021exploring, marx2021method, moses2022museum, liu2025spatial, tian2023expanding, liu2024spatiotemporal}.
ST-enabled studies have advanced our understanding of intra- and inter-tumoral heterogeneity \cite{li2022spatial,wang2023single,valdeolivas2024profiling}, the tumor microenvironment \cite{hwang2022single, khaliq2024spatial, hirz2023dissecting, peng2022spatial}, cellular interactions\cite{oliveira2025high, liu2025conserved,zohora2025cellnest,cang2023screening}, and the mechanisms driving cancer progression and resistance to therapy \cite{aung2025spatial,du2024integration,larroquette2022spatial,zhang2023spatial,xia2025deciphering}.
However, the high cost and technical demands of ST limit its adoption in cohort-scale studies \cite{moffitt2022emerging, tian2023expanding}. As an alternative, mounting evidence that morphological features in histology reflect underlying transcriptional states has motivated efforts to infer gene expression directly from hematoxylin and eosin (H\&E)-stained slides.

These efforts, collectively producing what is referred to as virtual ST, have rapidly evolved alongside broader advances in deep learning, and the underlying models can be grouped into three families: regression-based, bi-modal alignment-based, and generative-based. Regression-based methods directly predict gene expression from H\&E images. Building on the seminal ST-Net \cite{he2020integrating}, subsequent models have incorporated neighboring spot images, multi-scale context \cite{chung2024accurate, nonchev2025deepspot}, exemplar sets from reference datasets \cite{yang2023exemplar, YANG2024109966}, and textual gene descriptions \cite{Yan_Spatial_MICCAI2024}. Bi-modal alignment methods adapt the CLIP \cite{radford2021learning} paradigm to learn a joint morphology–expression embedding space for retrieval-based inference \cite{xie2023spatially, shi2024spatial, min2024multimodal, chen2025visual, han2025towards}. More recently, generative methods based on diffusion \cite{ho2020denoising} or flow matching \cite{lipman2022flow} have been introduced to model the joint distribution of gene expression rather than produce deterministic point estimates \cite{zhu2025diffusion, huang2025scalable}. Underlying all three families, the capacity of the image encoder to capture histomorphology has become a central determinant of performance, a role increasingly filled by pathology foundation models (PFMs) pretrained on large-scale H\&E data \cite{wang2022transformer,filiot2024phikonv2largepublicfeature,nechaev2024hibou,xiang2025vision,chen2024towards,lu2024visual,xu2024whole,vorontsov2023virchow,hoptimus1}.

Despite these advances, progress in virtual ST is increasingly constrained by inadequate benchmarking. Existing comparisons typically rely on small, heterogeneous datasets and inconsistent training and evaluation protocols, undermining the fairness of reported performance differences. Compounding this further, prior studies typically inherit each model's original image encoder rather than standardizing it across methods, entangling architectural innovations with encoder choice. These encoders differ substantially in pretraining domain, scale, and learning paradigm, and reported gains may therefore reflect feature extraction as much as model design. Equally problematic, evaluation has been confined to aggregate predictive accuracy, leaving the biological reliability of virtual ST and model robustness under realistic distribution shifts largely unexamined.

Two recent efforts have partially addressed these issues \cite{wang2025benchmarking, jaume2024hest}, but each leaves a complementary dimension uncontrolled. Wang et al. \cite{wang2025benchmarking} evaluated each method with its native image encoder, conflating architectural and encoder-specific contributions, and restricted their analysis to two early-generation ST datasets with a limited number of Visium samples. By contrast, HEST-1K \cite{jaume2024hest} assembled a large-scale collection of paired ST and histology samples, but its benchmarking focused on the raw correlation between PFM embeddings and gene expression using simple regressors, leaving the broader landscape of virtual ST architectures unexamined. No existing benchmark therefore enables controlled comparison of model architectures while standardizing the image encoder. In addition, both efforts evaluate models only at the level of aggregate gene-level correlation, without examining per-gene reliability, biological utility at the gene-set or cell-type level, robustness to domain shifts, or how performance scales with training data.

To bridge these gaps, we present $\ours$, a systematic benchmark for evaluating virtual ST under fair, controlled, and reproducible conditions, with the image encoder standardized across architecturally compatible models and all models evaluated under a unified implementation and evaluation framework. $\ours$ spans six cancer types and two ST platforms (Visium and Xenium), with each training dataset containing at least 30,000 spots and 15 slides to ensure sufficient statistical power. Within this benchmark, we re-implement and systematically profile $\nmodels$ models spanning regression-, bi-modal-, and generative-based families, all built on the same PFM image encoder when architecturally applicable. Evaluations span a wide range of dataset and PFM combinations, allowing us to identify the conditions that most strongly influence virtual ST quality. Beyond predictive accuracy, we characterize which genes and gene sets are reliably recoverable from histomorphology, revealing the molecular programs that virtual ST can and cannot capture. We further assess whether predicted profiles preserve the biological structure required for downstream analyses, including cell-type deconvolution \cite{li2022benchmarking, kleshchevnikov2022cell2location,miller2022reference,ma2022spatially,cable2022robust} and spatial domain identification \cite{kang2025benchmarking, hu2021spagcn, zhao2021spatial, dong2022deciphering}, and evaluate model robustness in cross-institutional, cross-tissue, and cross-platform settings. To support reproducibility and broad community adoption, the unified implementation of all $\nmodels$ models and the full benchmark are publicly released at \url{https://github.com/NEXGEM/STP-Bench}.

\Heading{Results}

\begin{figure*}
\centering
\includegraphics[width=0.97\textwidth]{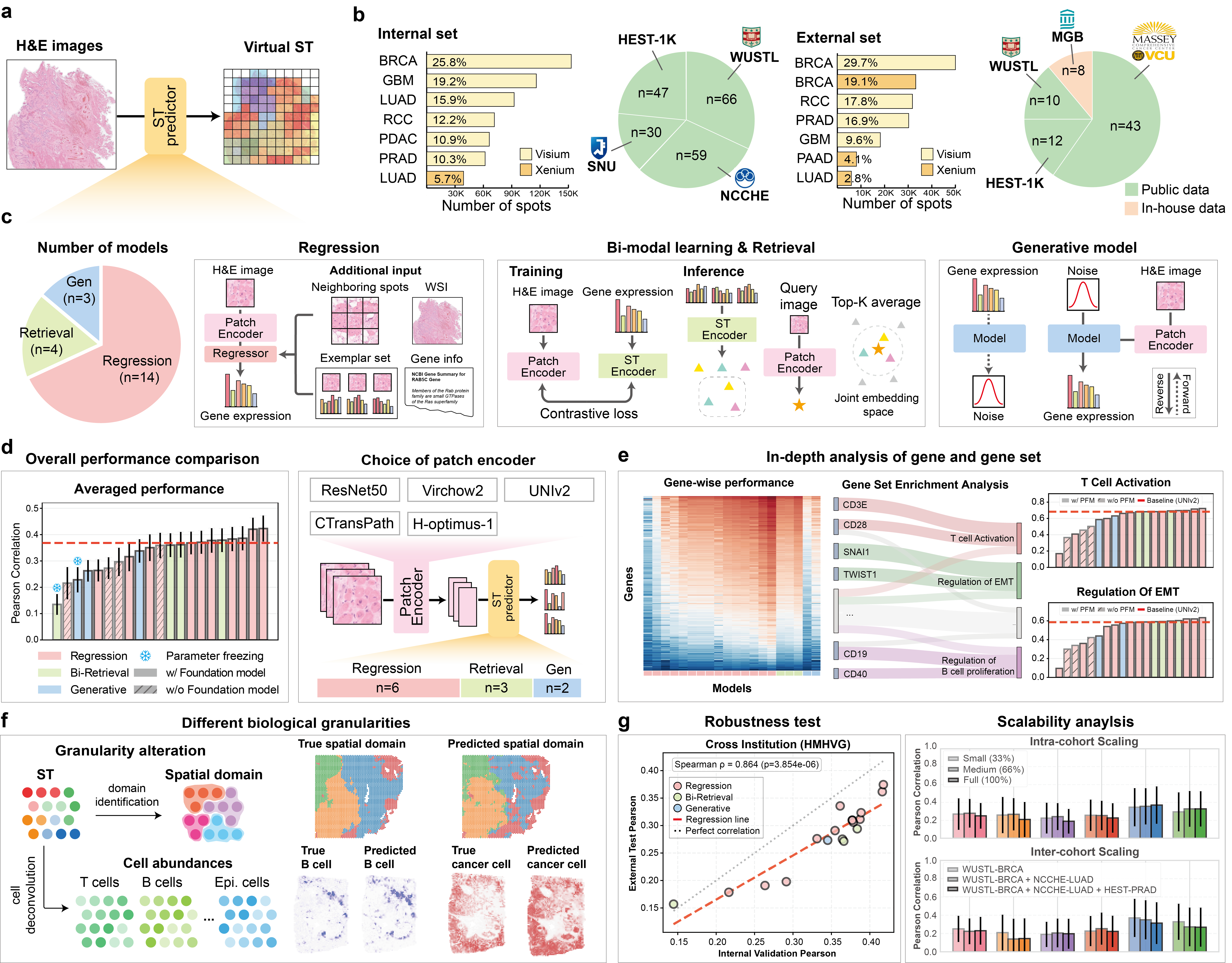}
\caption{\textbf{Overview of $\ours$ and benchmarked ST prediction models}.
\textbf{(a)} Virtual ST is generated by predicting spatial gene expression from H\&E-stained histology image patches.
\textbf{(b)} Distribution of the large-scale dataset that combines Visium and Xenium platforms, categorized into $\oursint$ (609,024 spots across 6 cancer types) and $\oursext$ (\nspotext\ spots across 6 cancer types) for training and testing, respectively.
\textbf{(c)} Workflows of the three methodological categories (a total of $\nmodels$ methods) evaluated on $\ours$: Regression models (14 models), Bi-modal learning \& Retrieval models (4 models), and Generative models (3 models). Diagrams depict the training and inference procedures for each category.
\textbf{(d)} Extensive evaluations within $\ours$. This includes overall predictive performance comparison and impact of different patch image encoders.
\textbf{(e)} In-depth analysis of gene and gene set level prediction quality, including a gene-wise performance comparison across all benchmarked models, gene set analysis linking representative marker genes to functional gene sets, and per gene set Pearson Correlation Coefficient (PCC) comparisons.
\textbf{(f)} Evaluation at different biological granularities, demonstrating the utility of virtual ST profiles for downstream tasks including spatial domain identification and cell type deconvolution.
\textbf{(g)} Robustness and scalability analyses, comprising a cross-institution correlation between internal validation and external test performance, and data scaling experiments under intra-cohort and inter-cohort settings.}
\label{fig:overview}
\end{figure*}

\heading{$\ours$: Unified, large-scale benchmark for virtual Spatial Transcriptomics} 

\noindent We established a large-scale benchmark dataset for virtual ST prediction from H\&E-stained tissue images, termed $\ours$, for both internal validation ($\oursint$) and external test validation \\ ($\oursext$) (\textbf{Extended Data Table~\ref{tab:dataset_composition}}). The dataset spans samples from six distinct cancer types and incorporates data generated using two ST acquisition platforms, Visium and Xenium (\textbf{Fig. \ref{fig:overview}b}). 
$\ours$ is formulated as a spot-level benchmark for virtual ST prediction, in which each spot aggregates transcripts from several neighboring cells and can be paired with an H\&E image patch centered on that spot.
$\oursint$ consists of \nspotint\ spots aggregated across \nslideint\ slides sourced from four data sources (HEST-1K\cite{jaume2024hest}, SNU \cite{sonpatki2026spatially}, NCCHE \cite{takano2024spatially}, and WUSTL\cite{mo2024tumour}). This cohort is designed primarily to benchmark virtual ST prediction performance across model architectures, target genes, and patch encoders.
$\oursext$ consists of \nspotext\ spots aggregated across \nslideext\ slides obtained from four data sources (HEST-1K, WUSTL, Massey\cite{bassiouni2023spatial}, and MGB). This cohort is used to externally validate models trained on $\oursint$ and to assess generalization performance across institutions, cancer types, and acquisition platforms. For unified evaluation, Visium datasets were processed by cropping 224 × 224-pixel H\&E image patches centered on each spot. For Xenium datasets, single-cell gene-expression profiles were first spatially pooled into Visium-scale pseudo-spots.
Details of each dataset and the preprocessing procedures are provided in \textbf{Online Methods}. 

\noindent We systematically evaluate $\nmodels$ models for virtual ST prediction, categorized into three distinct approaches (\textbf{Fig. \ref{fig:overview}c}): \textit{regression-based} (n=14), \textit{retrieval-based} (n=4), and \textit{generative} models (n=3) (\textbf{Extended Data Table~\ref{tab:st_gene_prediction_models}}). Regression-based approaches constitute the predominant class of methods, which treat the prediction of a gene expression panel as a multi-output regression problem based on H\&E image inputs. These models employ a patch encoder to compress the H\&E image patches cropped around each ST spot into a low-dimensional image embedding, followed by subsequent layers to map the image embeddings to gene expression values. While the original ST-Net focused only on localized morphology with image patches \cite{he2020integrating}, subsequent works have sought to additionally incorporate broader morphological context and complementary cues derived from external sources. Specifically, the morphological context ranges from neighboring regions all the way to whole-slide images to account for global morphological gradients \cite{pang2021leveraging, zeng2022spatial, chen2023spatial, chung2024accurate, nonchev2025deepspot, wang2024m2ort, wang2025m2ost}. The external cues center around either exemplar sets, which serve as a reference library of morphologically similar image patches paired with their corresponding ground-truth expression profiles \cite{yang2023exemplar, YANG2024109966}, or textual descriptions of genes for infusing prior biological knowledge \cite{Yan_Spatial_MICCAI2024}. We integrated fourteen regression-based models in our benchmark \cite{he2020integrating,rahaman2023breast,zeng2022spatial,yang2023exemplar,YANG2024109966,Yan_Spatial_MICCAI2024,chen2023spatial,chung2024accurate,nonchev2025deepspot,wang2024m2ort,wang2025m2ost,shulman2025path2space,mejia2023sepal,kumar2024histospace}.

\noindent Alternatively, retrieval-based approaches rely on CLIP\cite{radford2021learning}-style contrastive learning to align H\&E image patch inputs and corresponding gene expression inputs within a joint embedding space during training. Specifically, image patches are encoded into low-dimensional image embeddings via a patch encoder, while ST profiles are projected to the same latent space using a dedicated gene expression encoder, and the two modalities are aligned via a bidirectional contrastive loss. More recent variants further augment the gene expression branch with learnable spatial coordinate embeddings or lightweight Transformer-based spot encoders to capture spatially varying expression patterns beyond coordinate-free alignment \cite{shi2024spatial,min2024multimodal}, or reformulate transcriptomic profiles as rank-based gene "sentences" to mitigate batch effects and sequencing depth variability and enable large-scale vision--omics foundation modeling \cite{chen2025visual} . For predicting gene expression for a given image input, the models retrieve the top-k closest reference gene expression profiles from a reference database based on embedding similarity, and aggregate them via either uniform or similarity-weighted averaging to produce the final prediction. Four retrieval-based models are included in our benchmark \cite{xie2023spatially,shi2024spatial,min2024multimodal,chen2025visual}.

\noindent Benefitting from recent advances in generative models, another line of work has proposed reformulating ST prediction as a conditional generative modeling problem. Specifically, H\&E images are first encoded into latent embeddings, which act as a conditional variable for the subsequent generative process, instantiated via diffusion, flow matching, or masked generative modeling. The gene expression predictions are then obtained from the learned conditional model, with stochastic sampling enabling interval estimates. Three generative models are included in our evaluation pipeline \cite{zhu2025diffusion,huang2025scalable,huang2025stpath}.

\noindent To systematically compare these diverse modeling paradigms under fair conditions, $\ours$ is designed for comprehensive benchmarking of performance, robustness, and practical utility.
Our benchmarking framework consists of three major components: 
(1) Comprehensive performance evaluation with a systematic investigation of alternative image encoders with $\oursint$ (\textbf{Fig. \ref{fig:overview}d}); 
(2) In-depth multi-granularity analysis spanning gene-level expression, gene-set, cell composition, and spatial domain (\textbf{Fig. \ref{fig:overview}e-f}); and 
(3) Assessment of model robustness and scalability using $\oursext$ (\textbf{Fig. \ref{fig:overview}g}). Across the different evaluation scenarios, we use the Pearson correlation coefficient (PCC), which measures the linear correlation between predicted and observed expression values and is one of the most widely employed metrics for virtual ST evaluation\cite{he2020integrating,rahaman2023breast,zeng2022spatial,yang2023exemplar,YANG2024109966,Yan_Spatial_MICCAI2024,chen2023spatial,chung2024accurate,nonchev2025deepspot,wang2024m2ort,wang2025m2ost,shulman2025path2space,mejia2023sepal,kumar2024histospace}, and Structural Similarity Index Measure (SSIM), which additionally accounts for the spatial distribution of expression\cite{zhang2024inferring,almagro2025ai}, and the mean absolute error (MAE), which quantifies the average magnitude of deviation between predicted and observed expression values. Detailed definitions and computation procedures for all three metrics are provided in the \textbf{Online Methods}.

\begin{figure*}
\centering
\includegraphics[width=0.97\textwidth]{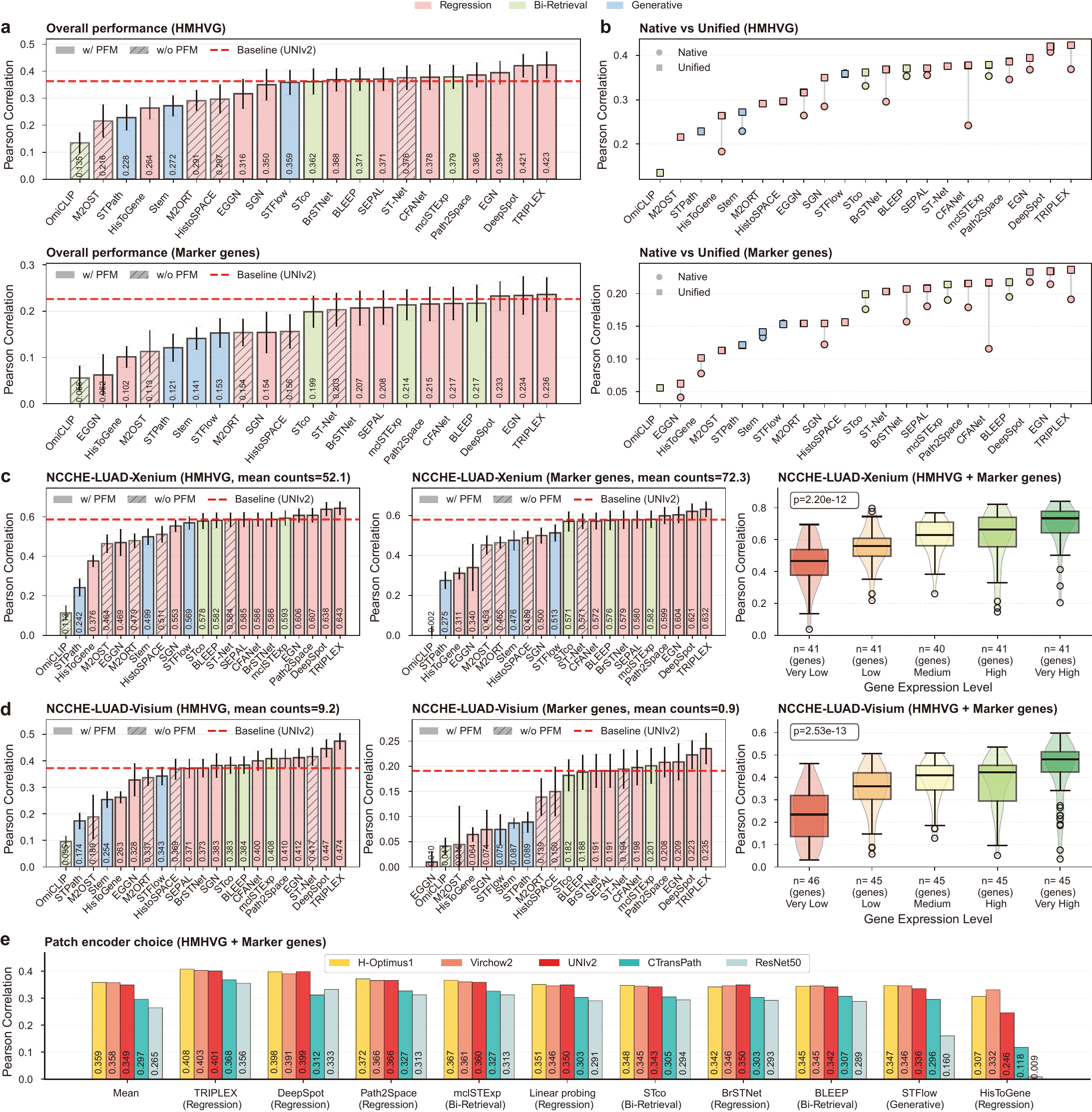}
\caption{\textbf{Performance benchmark of virtual ST baselines on $\oursint$}.
\textbf{(a)} The average PCC of virtual ST models evaluated on high-mean highly-variable genes (HMHVG) (top) and 16 marker genes (bottom), across 7 evaluation datasets. For each model, the patch features were extracted with UNIv2 (w/ PFM) unless a model required a specific patch feature extractor (w/o PFM, diagonal lines). The baseline (dashed line) represents the linear probing performance with UNIv2 image features. Error bars indicate the standard deviation across cross-validation folds.
\textbf{(b)} The comparison between virtual ST models when using UNIv2 feature extractor and the native feature extractors from respective original studies.
\textbf{(c-d)} Evaluation on representative NCCHE-LUAD-Xenium (c) and NCCHE-LUAD-Visium (d) datasets. Averaged gene counts are shown for each gene across datasets. The bar charts (left) rank the models based on their PCC performance on each specific dataset, with error bars indicating the standard deviation across cross-validation folds. The box plots (right) illustrate the distribution of prediction accuracy (PCC) across five gene expression levels (Very Low to Very High), with Kruskal-Wallis test results indicating statistical significance. 
\textbf{(e)} PCC for ten selected models that utilize different patch encoders on the union of HMHVGs and marker genes: H-Optimus1, Virchow2, UNIv2, CTransPath, and ResNet50.
}
\label{fig:OverallPerf}
\end{figure*} 

\heading{Performance benchmark on $\oursint$}

\noindent We benchmarked performance on $\oursint$ across two distinct gene sets: (1) 200 high-mean highly-variable genes (HMHVGs), selected based on jointly ranking genes by their mean expression and variability across slides; and (2) 16 key marker genes characterizing the tumor microenvironment (TME). The TME panel encompasses distinct functional lineages, including epithelial (\textit{KRT5}), stromal and vascular (\textit{ACTA2}, \textit{VIM}, \textit{PECAM1}), T-lymphocyte (\textit{CD3E}, \textit{CD4}, \textit{CD8A}, \textit{CXCR5}), B-lymphocyte (\textit{CD19}, \textit{CCR7}), myeloid lineage (\textit{CD68}, \textit{CD163}, \textit{ITGAX}, \textit{ITGAM}), and immune checkpoint (\textit{PDCD1}, \textit{CD274}) markers. This marker selection procedure is further detailed in the \textbf{Online Methods} section. 

\noindent Performance was assessed using k-fold cross-validation within each dataset, with the original fold assignments retained for the HEST-bench datasets and 5-fold patient-level cross-validation applied to all remaining datasets. For each slide, the evaluation metric was computed by comparing the ground-truth labels with the model predictions across ST spots. These values were first averaged across all samples within each fold and subsequently averaged across all folds. The mean performance for these categories across all datasets is summarized in \textbf{Fig. \ref{fig:OverallPerf}a}.
To ensure a fair comparison across different models, we standardized the patch encoders to a state-of-the-art pathology foundation model (PFM), UNIv2 \cite{chen2024towards}, whenever applicable. 
This standardization allows us to isolate and assess the effects of the architectural advantages of regression, retrieval, and generative frameworks on the downstream performance. 
Exceptions were made for models that were evaluated without PFM integration, namely OmiCLIP, M2ORT, M2OST, HistoSPACE, and ST-Net (denoted as \textit{w/o PFM} and hatched lines in \textbf{Fig. \ref{fig:OverallPerf}}), due to architectural incompatibility, use of a fixed pretrained encoder, or redundancy with other evaluated configurations; full details are provided in the Online Methods. We also report the linear probing performance with UNIv2 encoder as a baseline (dashed red line). Further details regarding specific model implementations can be found in the \textbf{Online Methods} section. 

\noindent Surprisingly, the overall performance in the internal evaluation indicates that most existing models failed to outperform the simple linear probing baseline with UNIv2 (\textbf{Fig. \ref{fig:OverallPerf}a}). In particular, most models without PFM fell below the baseline, along with some of the \textit{w/ PFM} models. Among the models that utilized PFMs, the regression-based approaches that integrate image features from multiple region-levels, TRIPLEX and DeepSpot, consistently surpassed this baseline. Other models, including Path2Space (non-linear regression with PFM features), EGN (exemplar-based), and mclSTExp (coordinate-aware bi-retrieval), only marginally surpassed the baseline. The evaluation revealed a substantial performance gap between gene categories. While HMHVGs were predicted with moderate accuracy (PCC $\approx 0.4$), marker genes showed remarkably lower predictability (PCC $\approx 0.2$) (\textbf{Fig. \ref{fig:OverallPerf}a}). Despite the difference, the model rankings remained largely consistent across both categories, indicating that models performing well on HMHVGs tended to exhibit similar comparative performance on marker genes. This trend was further supported by gene-wise stratified analyses, in which genes were grouped into five predictability bins based on their mean PCC across models (\textbf{Extended Data Fig. \ref{fig:Genewise_hmhvg}}). The relative model rankings remained largely stable across all bins, indicating that models with strong overall performance tended to remain superior regardless of gene predictability.

\noindent To further investigate the importance of the patch encoder on each method, we compared the performance of the models using the original patch encoders as implemented in the respective publications (\textit{Native}) against those using UNIv2 (\textit{Unified}) (\textbf{Fig. \ref{fig:OverallPerf}b}). 
We observe that the performance rankings between models using the original encoders are inconsistent with the rankings when using UNIv2. For instance, CFANet \cite{chen2023spatial}, which ranked 16th when using its native encoder, a simple linear projection of image patches into a fixed dimension, jumped to 6th when using UNIv2. We additionally observe that the performance gap was particularly pronounced in models utilizing encoders trained from scratch (CFANet, HisToGene) or pre-trained on non-pathology domains (BrSTNet). These results underscore the importance of leveraging powerful morphological representations learned from hundreds of millions of H\&E patches with PFM. Furthermore, these suggest that the evaluations in previous works, which did not account for the differences in underlying image backbones, are likely unfair comparisons and need to be taken cautiously.

\noindent To further investigate the sources of the performance gap between HMHVG and marker gene categories, we further conducted analyses at the individual dataset level (\textbf{Fig. \ref{fig:OverallPerf}c-d, Extended Data Fig. \ref{fig:Data_wise_performance}}). The analyses reveal that the gap is largely driven by data sparsity rather than morphological uncoupling alone. In the NCCHE-LUAD-Xenium dataset, which had substantially higher gene counts, both HMHVGs and marker genes were predicted with high accuracy across multiple models, including the UNIv2 baseline (PCC = 0.586 for HMHVGs and PCC = 0.579 for marker genes; \textbf{Fig. \ref{fig:OverallPerf}c}). By contrast, Visium-based datasets (e.g., NCCHE-LUAD-Visium) exhibited much lower gene counts, particularly for marker genes. In this setting, marker-gene prediction was substantially weaker, with models generally remaining in a low-performance range around the UNIv2 baseline (PCC = 0.191 for marker genes; \textbf{Fig. \ref{fig:OverallPerf}d}).
This trend was further supported by the stratified analysis of basal gene expression levels, where genes with higher basal expression generally showed better prediction performance across datasets (\textbf{Fig. \ref{fig:OverallPerf}c-d}, box plots). Prediction performance differed significantly across expression-level groups, with substantial effect sizes in both NCCHE-LUAD-Xenium (Kruskal-Wallis $H(4)=60.57$, $p=2.20\mathrm{e}{-12}$, $\epsilon^2=0.284$) and NCCHE-LUAD-Visium ($H(4)=65.04$, $p=2.53\mathrm{e}{-13}$, $\epsilon^2=0.276$). These results suggest that the quality of the underlying transcript detection plays a role in the model's predictive performance. 

\noindent Given the pivotal role of the patch encoder in capturing histological semantics \cite{campanella2025clinical, neidlinger2025benchmarking}, we assessed the predictive performance with the combined set of HMHVG and marker genes across ten different models, where each model was evaluated with five different image encoders: UNIv2 \cite{chen2024towards}, Virchow2 \cite{zimmermann2024virchow}, H-Optimus1 \cite{hoptimus1}, CTransPath \cite{wang2022transformer}, and ResNet50 \cite{he2016deep} (\textbf{Fig. \ref{fig:OverallPerf}e}). These encoders span a wide spectrum in model size, architecture, and pretraining data domain, scale, and diversity. UNIv2, Virchow2, and H-Optimus1 are ViT \cite{dosovitskiy2020image}-based models trained on large-scale in-house WSI datasets ranging from hundreds of thousands to over one million slides using DINOv2 \cite{oquab2023dinov2}, CTransPath was trained on a smaller collection of publicly available WSIs, and ResNet50 serves as an ImageNet\cite{deng2009imagenet}-pretrained convolutional baseline (\textbf{Extended Data Table~\ref{tab:pretrain_models}}). We also include linear probing based on each of the encoders as the reference baseline.

\noindent Similar to the observations in \textbf{Fig. \ref{fig:OverallPerf}b}, these experiments reaffirm the critical importance of the patch encoder. 
Specifically, large-scale pathology foundation models, namely UNIv2, H-Optimus1, and Virchow2, achieved the highest and broadly comparable performance across model choices. Despite being pretrained on pathology datasets, CTransPath showed only marginal improvements, which can be attributed to the small size of the pretraining dataset. The ImageNet-pretrained ResNet50 yielded the lowest performance independently of the model architecture, as expected due to the lack of domain-specific knowledge. 
Additionally, we observe that a subset of models, namely TRIPLEX, DeepSpot, Path2Space, and mclSTExp, consistently outperformed linear probing regardless of the encoder, underscoring the vital role of model-specific architectures. Specifically, TRIPLEX with the lowest-performing ResNet50 encoder still surpassed the performance of linear probing with the H-Optimus1 encoder. 
This underscores the combined importance of the architectural design of the ST prediction model and the vision encoder pretrained on large-scale and domain-specific pretraining datasets.
To assess whether these findings extend beyond correlation-based agreement, we further evaluated model performance using the Structural Similarity Index Measure (SSIM), which summarizes the structural agreement between virtual and measured spatial expression maps (\textbf{Extended Data Fig. \ref{fig:performance_SSIM}}). The SSIM-based results showed trends largely consistent with those obtained using PCC, supporting the overall conclusions regarding encoder importance and relative model performance.

\begin{figure*}
\centering
\includegraphics[width=0.95\textwidth]{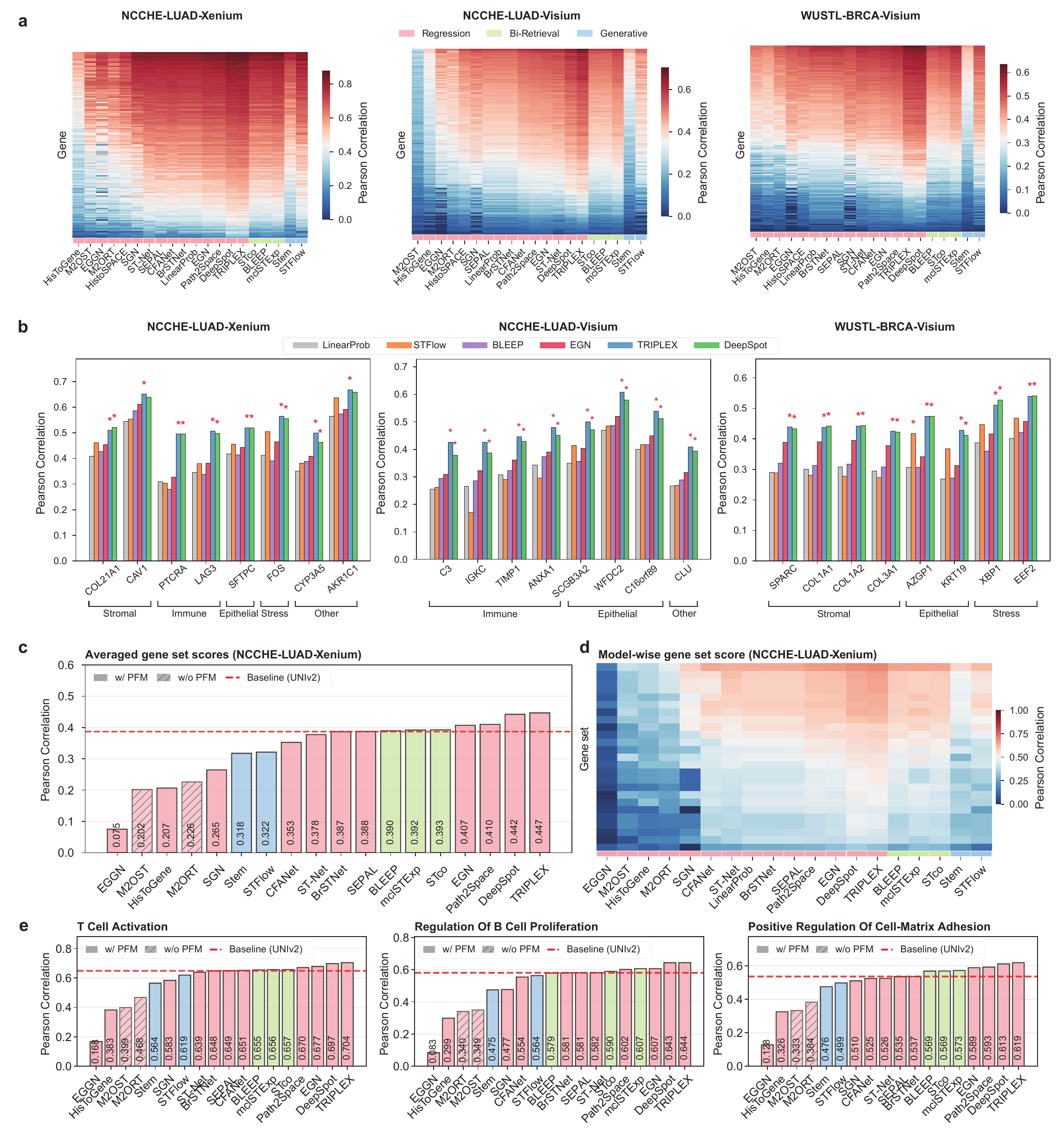}
\caption{\textbf{Individual gene and gene set analysis}.
\textbf{(a)} Gene-wise performance heatmap across 20 benchmarked methods. Each row and column represents individual genes (union of HMHVGs and marker genes) and prediction models, respectively, with the color bar below indicating the methodological category of each model. The color scale indicates the gene-wise PCC. Whenever applicable, UNIv2 was used as the patch encoder.
\textbf{(b)} Performance comparison for genes exhibiting high absolute performance (PCC $\ge 0.4$) and the highest predictive margin over the linear probing baseline. Bars represent the PCC for the linear baseline (LinearProb) and the selected models. Red asterisks indicate genes for which the selected model exceeded the baseline by more than 0.1.
\textbf{(c-e)} Gene set scoring prediction performance evaluated via singscore \cite{foroutan2018single} on NCCHE-LUAD-Xenium. Predicted gene expression profiles were used to compute per spot gene set scores using singscore, which were then compared against ground-truth scores derived from observed expression. \textbf{(c)} Bar chart showing averaged PCC across all evaluated gene sets. \textbf{(d)} Heatmap showing per gene set PCC across all benchmarked methods. \textbf{(e)} Bar charts showing gene set level results for representative pathways.
}
\label{fig:gene_analysis}
\end{figure*} 

\heading{Gene- and gene set-level analysis of virtual ST}

\noindent To understand the properties of individual genes or functional pathways that are well or poorly-predicted based on histomorphology, we further evaluated gene-wise prediction performance on $\oursint$ across 20 models. These included the linear probing baseline but excluded OmiCLIP and STPath, the two pretrained models from the previous section, which were evaluated in a zero-shot setting and were therefore not directly comparable to the fine-tuned models.
\noindent The gene-wise performance heatmap showed a smooth separation of high- and low-performing genes across models, with each gene predicted at a consistent accuracy level across methods, indicating that some genes are broadly predictable from histomorphology, whereas others remain difficult for nearly all models (\textbf{Fig.~\ref{fig:gene_analysis}a}). This gene-wise concordance was consistently observed across additional datasets (\textbf{Extended Data Fig.~\ref{fig:gene_geneset_add}a}), indicating that genes that are difficult (or easy) to predict for one model remain similarly challenging across diverse models.

\noindent Despite this shared gene-wise predictability pattern, the heatmap also showed model-dependent deviations in the transition between high- and low-performing genes, as reflected by the curved boundary of the performance gradient across methods rather than a strictly horizontal separation. This suggests that, although gene-level predictability is broadly conserved, architectural design choices can shift this predictability boundary and provide additional gains for specific genes or gene families. To quantify this model-dependent architectural gain, we selected six representative models spanning the major modeling paradigms evaluated in $\oursint$: the linear probing baseline, two regression-based models (TRIPLEX and DeepSpot), one bi-modal retrieval model (BLEEP), one generative model (STFlow), and one exemplar-guided regression model (EGN). For each gene, we computed the predictive margin as the difference between the highest PCC achieved by any representative model and the PCC of the linear probing baseline. We then focused on genes satisfying both high absolute predictive performance (PCC $\ge 0.4$) and a substantial gain over the baseline ($\Delta$PCC $\ge 0.1$), as detailed in the \textbf{Online Methods} section. Applying these criteria identified 8 genes in NCCHE-LUAD-Xenium, 85 genes in NCCHE-LUAD-Visium, and 37 genes in WUSTL-BRCA-Visium; for visualization, the top eight genes with the largest predictive margins in each cohort are shown in \textbf{Fig.~\ref{fig:gene_analysis}b}. Across cohorts, the largest gains were consistently achieved by DeepSpot and TRIPLEX, highlighting the advantage of architectures that integrate multi-scale morphological information. These observations motivated a gene-level characterization of the largest-margin genes to determine whether the architectural gains were concentrated in specific biological programs or spatial expression patterns.

\noindent A closer inspection of the genes with the largest predictive margin over the linear probing baselines reveals several coherent biological categories. First, stromal and ECM-remodeling programs, including fibrillar collagens and matrix regulators (\textit{COL1A1}, \textit{COL1A2}, \textit{COL3A1}, \textit{SPARC}, and \textit{COL21A1}), depend on meso- to macro-scale tissue patterns such as fibrosis gradients, matrix remodeling, and desmoplastic architecture\cite{winkler2020concepts,chakravarthy2018tgf}, which are not readily identifiable from isolated local patches. Next, immune- and inflammation-associated genes (\textit{IGKC}, \textit{C3}, \textit{LAG3}, \textit{TIMP1}, \textit{ANXA1}, and \textit{PTCRA}) reflect the spatial organization of lymphoid aggregates, peritumoral immune fronts, checkpoint-associated T-cell states, and diffuse inflammatory niches\cite{keren2018structured,schurch2020coordinated,pelka2021spatially,aggarwal2023lag}. Specialized epithelial differentiation markers, including lung secretory/alveolar genes (\textit{SFTPC}, \textit{SCGB3A2}, and \textit{C16orf89}) and breast luminal epithelial-associated genes (\textit{AZGP1} and \textit{KRT19}), are driven by the interplay between fine-grained cellular morphology and higher-order glandular or airway architecture\cite{sikkema2023integrated,kumar2023spatially}. Finally, stress and proliferation regulators (\textit{FOS}, \textit{XBP1}, and \textit{EEF2}) may benefit from models that contextualize localized cellular cues within broader hypoxic, reactive, or proliferative microenvironments\cite{hanahan2022hallmarks}.
Together, these patterns indicate that while models relying on local image patches may be sufficient for capturing broad epithelial structures, accurately decoding the complex, long-range morphological dependencies associated with the stroma and immune microenvironment requires architectures capable of synthesizing information across multiple spatial scales.

\noindent To further assess whether the gene-level predictive signals translate into biologically coherent functional representations, we evaluated gene set scoring prediction performance using a vectorized rank-based implementation following the singscore \cite{foroutan2018single}, which computes gene set activity independently for each spot and is well suited for efficient computation across the many spots, models, and cohorts in our benchmark. Genes were ranked within each spot, and gene set scores were obtained by averaging the ranks of member genes.
Specifically, predicted gene expression profiles were used to compute per spot gene set scores, which were then compared against ground-truth gene set scores derived from observed expression via PCC. As gene set scoring performance was substantially higher on NCCHE-LUAD-Xenium than on Visium-based cohorts, mirroring the gene-level prediction gap between platforms, we present results on NCCHE-LUAD-Xenium as the primary analysis; results for remaining cohorts are provided in \textbf{Extended Data Fig.~\ref{fig:gene_geneset_add}b}. The averaged performance across gene sets remained broadly consistent with gene-level evaluation trends (Spearman Correlation = 0.902), with TRIPLEX and DeepSpot again ranking among the top performers (\textbf{Fig.~\ref{fig:gene_analysis}c}), further corroborating that architectural advantages at the gene level translate to functional gene set representations. The per gene set heatmap (\textbf{Fig.~\ref{fig:gene_analysis}d}) reveals that well-predicted gene sets tend to be shared across methods, suggesting a common set of gene sets that are more readily recoverable from histology. At the same time, certain gene sets exhibit method-specific advantages, mirroring those gene-level patterns in which predictability varies across individual genes.

\noindent Examining individual gene sets revealed that both immune-related programs (\textit{T Cell Activation} and \textit{Regulation of B Cell Proliferation}) and a cancer-associated process (\textit{Positive Regulation Of Cell-Matrix Adhesion}) were recovered to a meaningful degree across top-performing models (\textbf{Fig.~\ref{fig:gene_analysis}e}). In particular, TRIPLEX and DeepSpot tended to perform more favorably on these gene sets, consistent with their strength in predicting immune- and stromal-associated genes at the individual gene level. This suggests that the multi-scale morphological reasoning employed by these models is especially effective for capturing spatially structured biological programs spanning both the immune and tumor microenvironmental contexts. 

\heading{Evaluating ST prediction models across multiple biological granularities}

\noindent Beyond spot-level gene expression accuracy, we assessed whether the virtual ST profiles preserve biologically meaningful structure at coarser levels of organization, namely cell type composition and spatial tissue architecture. We performed cell type deconvolution and spatial domain identification on both ground-truth and predicted expression matrices using the same six representative models selected for gene and gene-set analyses. The ground-truth-derived results served as the reference upper bound for each task.

\begin{figure*}
\centering
\includegraphics[width=0.97\textwidth]{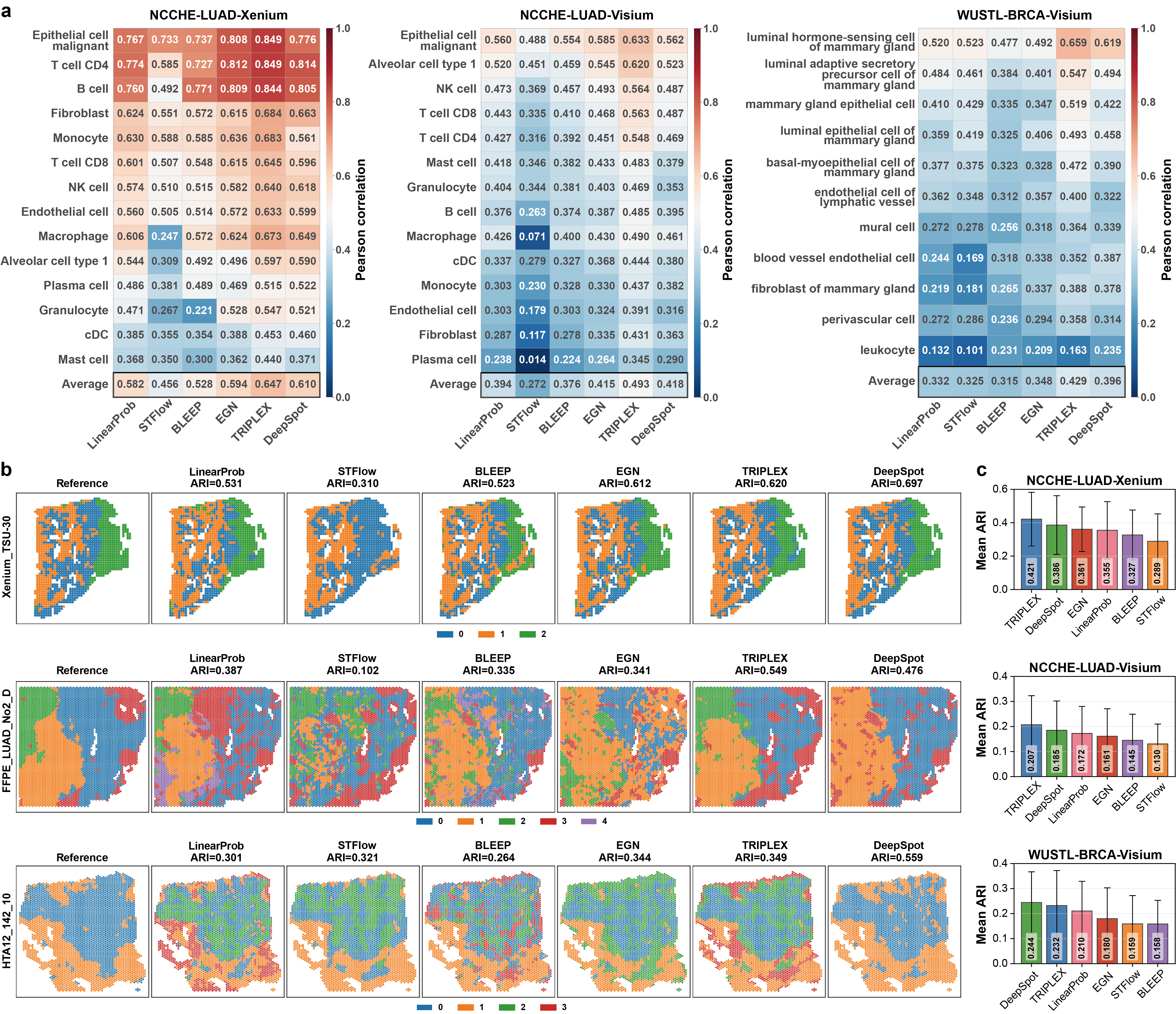}
\caption{\textbf{Evaluation at multiple biological granularities across three cohorts (NCCHE-LUAD-Xenium, NCCHE-LUAD-Visium, and WUSTL-BRCA-Visium)}. In all panels, reference annotations are derived by applying the respective downstream analysis directly to ground-truth gene expression profiles, serving as the upper bound for performance evaluation.
\textbf{(a)} Performance for cell type abundance estimated by Cell2location \cite{kleshchevnikov2022cell2location}. Heatmaps display per-cell-type PCC between predicted and ground-truth cell type abundance estimates for six representative models, with row-wise averages summarizing overall deconvolution accuracy.
\textbf{(b-c)}  Performance for spatial domain identification, where the domains are identified with SpaGCN \cite{hu2021spagcn}. Adjusted Rand Index (ARI) scores were computed between the spatial domains identified from the ground truth (Reference) and virtual ST from each method. \textbf{(b)} Visual comparison of representative tissue sections with reference and predicted spatial domain maps, where each color in the legend denotes a distinct spatial domain.  \textbf{(c)} Bar charts summarizing mean ARI scores across all test samples, with ARI annotated within each bar and sample-wise standard deviations indicated as error bars.
}
\label{fig:granularities}
\end{figure*}

\noindent For cell type deconvolution, we applied Cell2location\cite{kleshchevnikov2022cell2location} to virtual ST profiles from six models across three cohorts (\textbf{Fig.~\ref{fig:granularities}a}). The resulting cell-type abundance estimates were compared with the abundance derived from ground-truth expression. Across all three cohorts, TRIPLEX and DeepSpot consistently achieved the highest average correlation with the ground truth, consistent with their superior gene-level predictive performance. Abundances for major cell populations such as malignant epithelial cells and T cells were generally well-recovered, whereas rare or spatially diffuse populations, including plasma cells and mast cells, exhibited lower correlations. 

\noindent For spatial domain identification, we clustered predicted gene expression profiles using SpaGCN\cite{hu2021spagcn} and evaluated the resulting spatial partitions against reference domain annotations using the Adjusted Rand Index (ARI) (\textbf{Fig.~\ref{fig:granularities}b-c}). 
Across all three cohorts, DeepSpot and TRIPLEX consistently ranked among the top-performing models, with TRIPLEX achieving the highest mean ARI on NCCHE-LUAD-Xenium and NCCHE-LUAD-Visium, and DeepSpot leading on WUSTL-BRCA-Visium. In contrast, STFlow exhibited the lowest ARI across all cohorts, consistent with its relatively weaker gene-level performance. 

\noindent Notably, both downstream tasks yielded substantially higher performance for Xenium-based than Visium-based cohorts, mirroring the gene-expression prediction gap between platforms observed previously. 
This concordance suggests that improvements in gene-level prediction accuracy directly translate to downstream biological inference, and that transcript-level data quality constitutes a key determinant of virtual ST utility beyond individual gene metrics. 

\begin{figure*}
\centering
\includegraphics[width=0.97\textwidth]{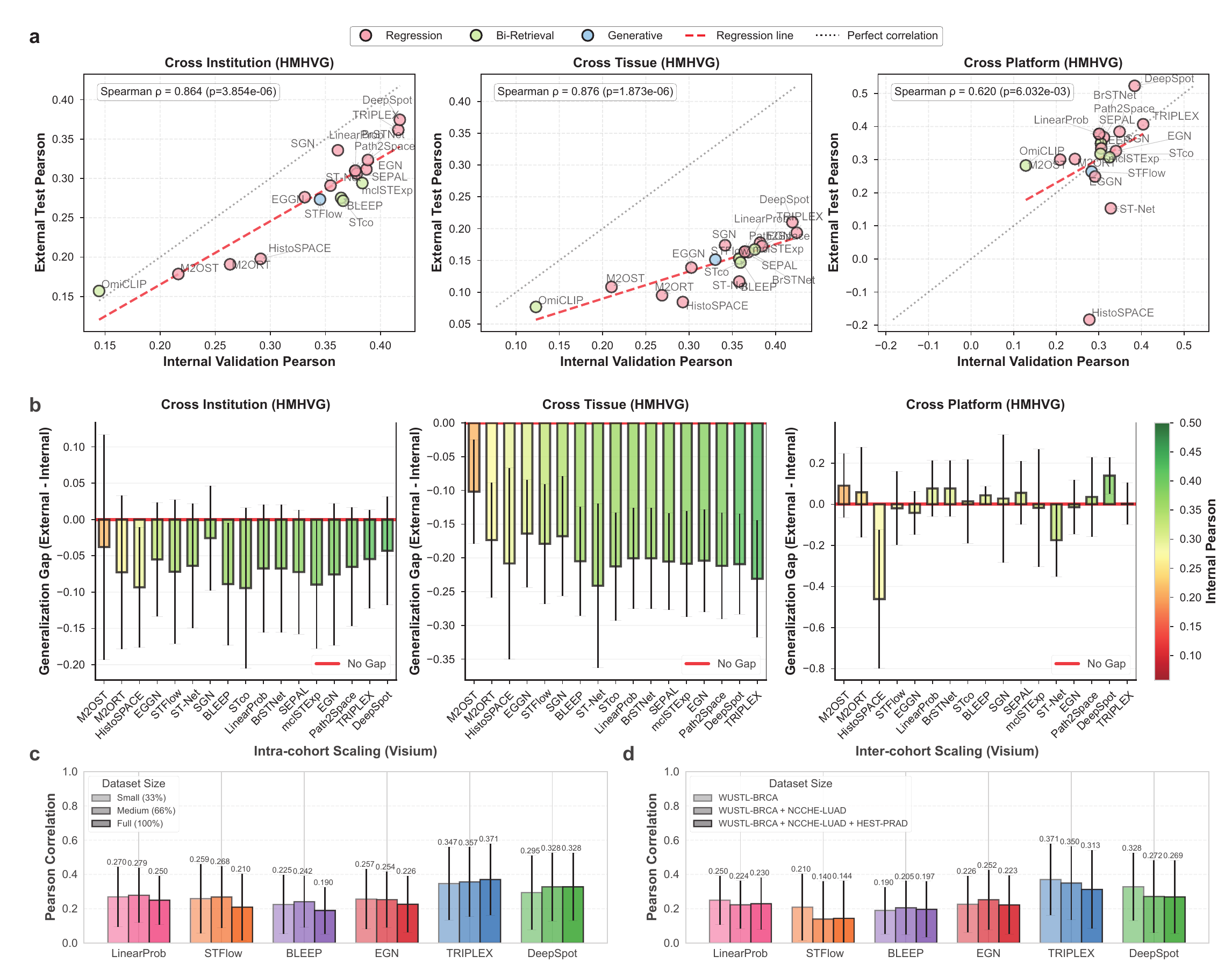}
\caption{\textbf{Robustness and Scalability Analysis}.
\textbf{(a)} Scatter plots illustrating the correlation between performance on $\oursint$ (x-axis) and $\oursext$ (y-axis) under three generalization scenarios: Cross-Institution (left), Cross-Tissue (center), and Cross-Platform (right). The displayed external performances are averaged across all dataset pairs (i.e., specific train-test data combinations) within each scenario. To assess the consistency of the model rankings, the Spearman correlation coefficient was used. Models are color-coded by category (Regression, Bi-Retrieval, Generative). The black dotted line indicates equal internal and external performance. The red dotted line indicates the fitted linear regression model to quantify the relationship between internal and external performance. Whenever applicable, UNIv2 was used as the patch encoder.
\textbf{(b)} Bar charts quantifying the generalization gap between the $\oursext$ and $\oursint$ performance for each model across the same scenarios as (a): Cross-Institution (left), Cross-Tissue (center), and Cross-Platform (right). The ``No Gap'' line (red) indicates no discrepancy between internal and external test performance. The models are ordered by the increasing PCC on $\oursint$, also represented by the color code. Error bars represent standard deviation of the performance differences across all dataset pairs.
\textbf{(c)} Bar chart showing the PCC performance of the selected models (LinearProb, STFlow, BLEEP, EGN, TRIPLEX, DeepSpot) when trained on subsets of increasing size (Small: 33\%, Medium: 66\%, Full) within the same dataset (WUSTL-BRCA-Visium). The error bars represent the standard deviation across all the samples in each cohort.
\textbf{(d)} Bar chart showing the impact of expanding training data across multiple cancer types (BRCA only vs. BRCA+LUAD vs. BRCA+LUAD+PRAD) on model performance. 
}
\label{fig:Generalization}
\end{figure*} 

\heading{Generalization and data scaling properties of ST prediction models}

\noindent To assess the robustness of ST prediction models against the common real-world domain shifts between training and testing datasets, we evaluated their performance across three distinct generalization scenarios: cross-institution, cross-tissue, and cross-platform transfers (\textbf{Fig. \ref{fig:Generalization}}). In the \textit{cross-institution} scenario, we evaluated model resilience to batch effects, such as technical artifacts from slide preparation protocols, by utilizing datasets of the same cancer type and ST platform collected from several different sites. This was extended to \textit{cross-tissue} transfer, which challenged the models to generalize across different anatomical sites, thereby testing whether organ-specific morphological features remain predictive in biologically distinct contexts. 
Finally, we conducted \textit{cross-platform} transfer to assess the model's ability to bridge the gap between different ST acquisition approaches, such as Visium and Xenium.
All of the generalization experiments were conducted with UNIv2 patch encoder whenever applicable.

\noindent First, we examined the correlation between the performance on internal benchmarks ($\oursint$) and external test dataset ($\oursext$) across these scenarios (\textbf{Fig. \ref{fig:Generalization}a}). We consistently observed a strong rank preservation of model performance (Spearman $\rho > 0.86$ and $p < 10^{-5}$ for both institution and tissue shifts), indicating that the architectural superiority established in internal benchmarks is generalizable. However, the absolute performance varied significantly depending on the nature of the domain shift (\textbf{Fig. \ref{fig:Generalization}b}). Cross-institutional shifts resulted in a moderate decline in performance, with the generalization gaps ranging from approximately $-0.05$ to $-0.10$ in PCC (\textbf{Fig. \ref{fig:Generalization}b}, left). This suggests that while technical variations do impact model accuracy, the morphological features remain sufficiently consistent for the model to maintain predictive utility. In stark contrast, cross-tissue generalization was challenging as evidenced by the sharper decline in performance, with the generalization gaps reaching $-0.20$ to $-0.35$ (\textbf{Fig. \ref{fig:Generalization}b}, middle).  
This indicates that biological heterogeneity due to different cancer morphologies presents far greater obstacles than technical variability. Interestingly, cross-platform transfer (Visium $\to$ Xenium) showed negligible performance loss, or even slight improvements in some cases (\textbf{Fig. \ref{fig:Generalization}b}, right). This likely reflects the higher sensitivity and transcript counts from Xenium dataset compensating for the domain shift.
\noindent To further investigate whether the observed generalization patterns depend on the choice of PFM, we conducted additional external test experiments by systematically replacing the patch encoder (\textbf{Extended Data Fig. \ref{fig:performance_corr}-\ref{fig:generalization_gap}}). 

\noindent Across cross-institutional transfers, all PFMs demonstrated strong rank preservation between internal and external performance, with consistently high correlation coefficients. This indicates that institutional batch effects do not substantially disrupt the relative architectural advantages, regardless of the underlying PFM. In contrast, cross-tissue transfer revealed clear differences across PFMs. While Virchow2 and H-Optimus1, similar to UNIv2, maintained moderate-to-strong correlations, ResNet and CTransPath exhibited markedly lower correlation coefficients, accompanied by noticeable shifts in model ranking. Notably, the PFMs that previously demonstrated superior patch-level performance (\textbf{Fig. 2e}) also showed better rank preservation under tissue shifts. This suggests that stronger morphological representation learning at the encoder level may contribute to improved cross-domain robustness. Finally, cross-platform transfer (Visium → Xenium) remained the most unstable scenario. Across all PFMs, both performance gains and declines were observed in a model-specific manner, without a consistent directional trend. This reinforces our earlier observation that cross-platform evaluation constitutes the most unpredictable external test setting.

\noindent Next, we investigated whether increasing the scale of training data translates to better generalization performance along two common real-world scenarios, \textit{intra-cohort scaling} and \textit{inter-cohort scaling}.
In \textbf{intra-cohort scaling}, we used increasing subsets of the spatial spots (33\%, 66\%, and 100\%) in the WUSTL-BRCA-Visium dataset, which contains the largest number of spatial spots, to evaluate the impact of data volume within a single domain. In \textbf{inter-cohort scaling}, we progressively merged multiple cohorts to increase diversity, starting with WUSTL-BRCA-Visium, and sequentially adding NCCHE-LUAD-Visium and HEST-PRAD-Visium, which previously demonstrated the highest internal validation performance. To ensure consistent evaluation across all scaling experiments, we used the independent Massey-BRCA-Visium test set and restricted our analyses to 30 HMHVGs common to all three training cohorts.
For this analysis, we focused on six models: the three top-performing regression models in the internal test (\textbf{Fig. \ref{fig:OverallPerf}a}), one representative from each of the generative (STFlow) and bi-modal retrieval (BLEEP) categories, and the linear probing baseline. Contrary to the ``scaling laws'' observed in foundation models\cite{zhai2022scaling}, increasing the training data volume within the same dataset (Intra-cohort scaling) yielded only marginal performance gains, suggesting that current models rapidly reach a saturation point regarding the information extractable from H\&E images (\textbf{Fig. \ref{fig:Generalization}c}). More critically, expanding data diversity by pooling distinct cancer types (Inter-cohort scaling) often led to negative transfer (\textbf{Fig. \ref{fig:Generalization}d}). Notably, as distinct tissue types were sequentially added (BRCA $\to$ +LUAD $\to$ +PRAD) to the training set, the performance on the target tissue mostly decreased. This indicates that the highly tissue-specific nature of the morphological features governing gene expression, and thus a simple aggregation of the heterogeneous sources could be detrimental to the predictive performance. A similar pattern was observed when the scaling experiments were repeated using other pathology foundation models (PFMs) as patch encoders, indicating that these trends were largely robust to the choice of PFM (\textbf{Extended Data Fig. \ref{fig:scale_pfm}}).

\Heading{Discussion}

In this work, we present $\ours$, a comprehensive and standardized benchmark for virtual spatial transcriptomics (ST) that addresses critical limitations of prior evaluation frameworks. By systematically evaluating 21 models across large-scale, multi-platform datasets under unified experimental conditions, we provide a rigorous and reproducible basis for assessing the state of the field.
A key finding of this study is that most existing models fail to outperform a simple linear probing baseline when a performant pathology foundation model (PFM) is used as the common image encoder. This challenges the prevailing assumption that architectural complexity alone drives predictive performance in virtual ST. Rather, our results demonstrate that the choice of patch encoder is a dominant factor, since replacing native encoders with a unified PFM substantially altered model rankings. This suggests that performance discrepancies reported in prior works may reflect differences in image representation rather than architectural innovation. This underscores the importance of disentangling image encoder choice from model design when assessing predictive performance and calls for greater transparency in how image backbones are selected and reported.   

Beyond overall performance, our gene-level analyses revealed a striking degree of concordance across models: genes that were difficult to predict for one model were consistently difficult for all others. This universal consistency suggests that prediction accuracy is governed more by the inherent correlation between morphology and underlying gene expression and less by the model architecture. Nevertheless, architectures capable of integrating multi-scale morphological information (most notably TRIPLEX \cite{chung2024accurate} and DeepSpot \cite{nonchev2025deepspot}) demonstrated the largest performance gains over the linear probing baseline for specific gene categories. Genes encoding stromal and ECM-remodeling programs, immune and inflammation-associated markers, and specialized epithelial differentiation factors benefited most from multi-scale reasoning, reflecting their dependence on meso- to macro-scale tissue organization that cannot be captured from isolated local patches alone. These gene-level advantages further translated to the gene set level: both immune-related programs and tumor microenvironment-associated processes such as cell-matrix adhesion were recovered more faithfully by multi-scale models, with model rankings remaining consistent across gene- and gene set-level evaluations.

Extending beyond gene-level accuracy, we further assessed whether predictive signals translate to coarser levels of biological organization through cell type deconvolution and spatial domain identification. Model rankings were largely preserved across both downstream tasks, mirroring the trends observed in gene-level evaluations. Notably, performance on both tasks was substantially higher for NCCHE-LUAD-Xenium than for Visium-based cohorts, directly reflecting the gene expression prediction gap between platforms. This concordance  suggests that the quality of the underlying gene expression data---in terms of transcript counts and signal-to-noise ratio---constitutes a shared determinant of performance across multiple granularities of biological analysis, and that improvements in data quality translate consistently from gene-level prediction to tissue-level biological inference.

Our robustness analyses revealed that internal model rankings were largely preserved under cross-institution and cross-tissue transfer, providing practical guidance for model selection in deployment settings. However,  cross-tissue generalization incurred substantially larger performance degradation than cross-institution transfer, underscoring the highly tissue-specific nature of morphology-to-expression mappings. Counterintuitively,  expanding training data across multiple cancer types through inter-cohort scaling often led to negative transfer, suggesting that simple data aggregation is insufficient---and potentially detrimental---when tissue types differ substantially. These observations point to the need for more principled approaches to multi-tissue training, such as domain-adaptive or tissue-aware learning strategies.

Several limitations of this study merit consideration. First, although $\ours$ represents the largest and most diverse benchmark for virtual ST to date, it remains constrained to six cancer types and two ST platforms.  Extending the benchmark to additional tumor types, tissue contexts, and emerging sequencing platforms will be important for broader generalizability. Second, our gene-level analyses were restricted to the high-mean highly-variable genes (HMHVGs) and 16 TME marker genes, and systematic investigation of lowly expressed or functionally diverse genes remains an open direction. 
Third, the evaluations are performed at the spot-level, where each spot aggregates transcripts from multiple cells in the neighborhood. With the increasing availability of single-cell ST datasets and virtual ST models, future benchmarking efforts may extend these principles to evaluations at single-cell resolution.

Taken together, $\ours$ establishes a rigorous foundation for evaluating virtual ST methods and provides several actionable insights: the importance of the patch encoder over architectural design, the existence of a universal gene-level performance ceiling tied to morphological informativeness, and the platform-dependence of biological prediction limits. We anticipate that $\ours$ will serve as a shared reference for virtual ST, enabling more principled model development and fairer benchmarking as the field continues to mature.

\clearpage
\Heading{Online Methods}

\heading{Benchmark dataset curation and composition}

\noindent To establish a robust framework for evaluating histopathology-based virtual spatial transcriptomics (ST) models, we curated the $\ours$ dataset, which is categorized into \oursint\ (model training and internal validation) and \oursext\ (independent validation) cohorts (Extended Data Table~\ref{tab:dataset_composition}). 

\noindent \textbf{\oursint}:
\noindent While existing resources such as the HEST-1K dataset \cite{jaume2024hest} provide a valuable foundation for ST research, many individual datasets within such collections lack the requisite scale to effectively train high-capacity deep learning models for specific cancer types. To ensure sufficient morphological diversity and statistical power for model convergence, we established a minimum inclusion threshold of $30{,}000$ spots and $15$ tissue slides per training set.
\noindent Based on these criteria, we selected the Prostate Adenocarcinoma (PRAD; 2 patients, 23 slides, 62,710 spots) and Clear Cell Renal Cell Carcinoma (CCRCC; 24 patients, 24 slides, 74,220 spots) datasets from the HEST-bench collection. To further enhance the breadth and depth of our training data, we incorporated additional cohorts from NCCHE \cite{takano2024spatially} (LUAD-Visium; 30 patients, 43 slides, 96,954 spots and LUAD-Xenium; 16 patients, 16 slides, 34,519 spots), WUSTL \cite{mo2024tumour} (BRCA; 21 patients, 47 slides, 157,138 spots and PDAC; 14 patients, 19 slides, 66,447 spots), and SNU \cite{sonpatki2026spatially} (GBM; 17 patients, 30 slides, 117,036 spots). In total, the internal cohort comprises seven distinct training sets spanning six cancer types, \nslideint\ tissue slides, and \nspotint\ capture spots. These datasets represent two major ST platforms, 10x Genomics Visium and Xenium, providing a multi-platform basis for model development.

\noindent \textbf{\oursext}:
\noindent To rigorously evaluate the generalizability of the models across different clinical environments and sample preparation protocols, we curated an external cohort that is strictly independent of the internal training data. A primary requirement for this cohort was the comprehensive coverage of all cancer types present in the internal sets to facilitate direct cross-cohort performance comparisons. This external benchmark was compiled by aggregating data from three public sources: HEST-1K (IDC; 4 patients, 4 slides, 35,536 spots, LUAD; 2 patients, 2 slides, 5,206 spots, GBM; 2 patients, 3 slides, 17,763 spots, and PAAD; 3 patients, 3 slides, 7,571 spots), WUSTL (RCC; 10 patients, 10 slides, 33,090 spots), and Massey \cite{bassiouni2023spatial} (TNBC; 22 patients, 43 slides, 55,227 spots), along with an internal source from Mass General Brigham (MGB; PRAD, 8 patients, 8 slides, 31,438 spots). In total, the external cohort comprises \nslideext\ slides and  \nspotext\ capture spots across all six cancer types. By maintaining complete separation between the training and validation cohorts, we ensure that the benchmark provides a blinded assessment of a model's ability to generalize to unseen patient populations and technical variations.

\heading{Data preprocessing}

\noindent To standardize inputs across heterogeneous datasets, our preprocessing pipeline was designed with platform-specific handling for Visium and Xenium data, while producing a unified output format compatible with the HEST library \cite{jaume2024hest}. This compatibility ensures that all processed datasets, regardless of their original platform, can be seamlessly loaded and managed through a common data structure (HESTData), facilitating reproducible benchmarking and consistent downstream analysis.

\noindent \textbf{Visium preprocessing:}
For Visium data, we utilized the predefined spot barcodes and their associated spatial coordinates provided in the raw data. Using these coordinates, whole-slide images were cropped into spot-centered patches of $224 \times 224$ pixels at a target resolution of $0.5\,\mu$m/pixel. The gene expression matrices (AnnData) were then subset and reindexed to strictly match the barcodes of successfully extracted patches, ensuring exact one-to-one cross-modal pairing. Datasets already distributed in HEST-compatible format (e.g., HEST-bench collections) were used directly without additional processing.

\noindent \textbf{Xenium preprocessing:}
For Xenium data, which provides single-molecule transcript coordinates rather than predefined spot grids, an additional binning step was required to construct ST pseudo-spots. Specifically, individual transcripts were aggregated into spatially regular bins of $100\,\mu$m $\times$ $100\,\mu$m, comparable to the inter-spot spacing of the Visium platform, to generate pseudo-spot expression matrices. The corresponding H\&E images, previously registered to DAPI-based morphology images via rigid transformation (e.g., using Xenium Explorer), were loaded alongside these pre-aligned pseudo-spot coordinates. Guided by the binned spatial positions, image patches of $224 \times 224$ pixels were extracted at the same $0.5\,\mu$m/pixel resolution as Visium. The resulting expression matrices and image patches were then packaged into the HEST-compatible format, enabling unified downstream processing across both platforms.

\noindent \textbf{Gene expression normalization:}
Following patch extraction and cross-modal alignment on both platforms, the gene expression matrices were subset to the target gene panel. A standard $\log(1+x)$ transformation (log1p) was applied to stabilize variance and reduce the outsized influence of highly expressed genes.

\heading{Benchmarking models}

\noindent We conducted a unified benchmarking and implementation framework encompassing 21 representative models for predicting spatial transcriptomic (ST) profiles directly from hematoxylin and eosin (H\&E)–stained histology images. All models were reimplemented and evaluated under a standardized framework. Models requiring task-specific training were trained under a unified configuration, whereas pretrained zero-shot models (OmiCLIP and STPath) were evaluated without fine-tuning. Again, to further ensure fair comparison, the image encoder of each model was replaced with a common pathology foundation model (PFM), eliminating performance differences attributable to encoder pretraining. Models whose architectures are structurally incompatible with a drop-in PFM replacement (e.g. HistoSPACE, M2ORT, M2OST) retained their original encoder designs. ST-Net was also kept in its original form, as substituting its encoder with a frozen PFM would reduce the model to a linear probing baseline. In our experiments, we adopted UNIv2 \cite{chen2024towards} as the default PFM. To enforce non-negativity of predicted gene expression values, a Softplus activation function was applied to the output layer of all models. Exceptions were made for retrieval-based models (BLEEP, STco, and mclSTExp), whose predictions are derived by aggregating retrieved ground-truth expression profiles and thus require no explicit non-negativity constraint, and for generative models (STFlow and Stem), whose outputs were instead clamped to zero at inference time.

\noindent \textbf{ST-Net}:
Following He et al. \cite{he2020integrating}, we used a pretrained DenseNet-121 as the image encoder, which takes a $224 \times 224$ image and produces a 1,024-dimensional feature vector. A single linear layer then maps this representation to the expression levels of target genes.

\noindent \textbf{BrSTNet}:
BrSTNet \cite{rahaman2023breast} extends the ST-Net by augmenting the single-head regression architecture with an auxiliary prediction head. As in ST-Net, each $224 \times 224$ image patch is encoded by a CNN backbone and mapped to the expression levels of the target gene set through a main linear head. BrSTNet additionally introduces a secondary head that predicts the remaining genes outside the target panel, trained jointly via a weighted loss ($\mathcal{L} = \mathcal{L}_{\text{main}} + \lambda \cdot \mathcal{L}_{\text{aux}}$), where the auxiliary objective acts as a regularizer to improve generalization. The original work evaluated multiple backbone architectures (ResNet-101, Inception-v3, EfficientNet, ViT), finding EfficientNet-b0 to perform most favourably. In our implementation, BrSTNet used PFM embeddings as input to two linear prediction heads without additional hidden layers or dropout. The auxiliary head predicted all non-target genes in the dataset-specific gene universe, and the auxiliary loss weight was set to $\lambda=1.0$.

\noindent \textbf{HisToGene}:
HisToGene \cite{pang2021leveraging} adopted a Vision Transformer (ViT) model for predicting spatial gene expression from histology. It uses a compact Transformer encoder with 4 layers, 16 attention heads, and a dropout rate of 0.1. In the original model, each H\&E image patch is flattened and linearly projected into a latent image embedding, which is then combined with learnable spatial coordinate embeddings. We adapted HisToGene to use pretrained PFM features by extracting spot-level image embeddings using PFMs and feeding these embeddings into the Transformer encoder.

\noindent \textbf{EGN}:
EGN \cite{yang2023exemplar} implements an exemplar-guided architecture for spatial gene expression prediction. The original framework employs a two-stage pipeline: first, a StyleGAN-based autoencoder is trained to learn a disentangled global view of each tissue patch, which is then used for unsupervised exemplar retrieval. Second, a ViT backbone processes the target patch, with Exemplar Bridging (EB) blocks interleaved between transformer layers. These EB blocks incorporate information from retrieved exemplar spots, including their global views and measured gene expression profiles, to condition the target representation on similar reference spots. 
For a fair comparison, we replaced only the global-view extractor and exemplar retrieval features with PFM embeddings. Exemplars were retrieved from the training fold using L1 distance in the PFM embedding space, and the top 9 exemplar spots, together with their measured expression profiles, were passed to the EB blocks. The target patch was still processed by the ViT backbone, while EB blocks injected target and exemplar PFM embeddings into the transformer representation. We used the original EGN transformer configuration with 16 layers, 16 attention heads, hidden dimension 1024, patch size $32$, and EB blocks inserted every two transformer layers.

\noindent \textbf{EGGN}:
EGGN \cite{YANG2024109966} extends the exemplar-guided framework by introducing explicit spatial and exemplar-graph reasoning. While the original model uses an ImageNet-1K pretrained ResNet for image feature extraction, we used PFM embeddings as window-node features and for exemplar retrieval. EGGN constructs a heterogeneous graph that captures window--window, exemplar--exemplar, and exemplar--window relationships. GraphSAGE layers and Graph Exemplar Bridging (GEB) blocks are then used to propagate information across this graph, and gene expression is predicted from the exemplar-refined window node features using a CSRA-style attention-based readout.
For graph construction, window--window edges were built using a spatial radius graph with radius $\sqrt{2}$ and at most five neighbors, while exemplar--exemplar edges were built with a KNN graph ($k=3$). Exemplar--window edges connected each window node to its retrieved exemplar nodes, where exemplars were selected from the training fold using L1 distance in the PFM embedding space. In our implementation, each window used the top 9 exemplars. We used four heterogeneous graph layers with hidden dimension 512, and trained the model with an MSE loss augmented by a correlation loss term weighted by 0.5.

\noindent \textbf{CFANet}:
CFANet \cite{chen2023spatial} predicts spatial gene expression using stacked coarse and fine attention network (CFAN) blocks that jointly model global and local visual context at multiple spatial scales. Each $224 \times 224$ spot image is divided into window tokens. Global context is captured by the Sparse Adaptive Global Area Attention (SAGA) module, which applies sparse attention over the top-$k$ most informative regions, whereas local context is modeled by the Multi-scale Dynamic Local Window Attention (MDLA) module, which uses dynamic convolution-based multi-scale intra-window attention. The resulting global and local representations are fused and passed to a linear regression head for gene expression prediction. We replaced the raw image tokenization with PFM embeddings: each spot-level PFM embedding was linearly projected into a $7 \times 7$ set of 1024-dimensional latent tokens and reshaped into a pseudo feature map compatible with the CFAN encoder. We used the CFA-small configuration with stage depths $[4,4,18,4]$ and embedding dimensions $[64,128,256,512]$.

\noindent \textbf{SEPAL}:
SEPAL \cite{mejia2023sepal} combines local patch-level prediction with graph-based spatial refinement. It first obtains initial spot-level gene expression predictions from local image embeddings. A spatial graph is then constructed over spots, where node embeddings include image embeddings and positional embeddings. These node features are processed by a MLP, followed by sequential GCNConv layers with ReLU activations and SAGPooling for graph-level aggregation. The GNN branch learns a residual correction, which is added to the initial local prediction to obtain the final gene expression estimate. 
Under the same evaluation protocol, initial local predictions were obtained by passing each PFM embedding through a spot-wise MLP regressor with hidden dimensions $[1024,512]$. These predictions were stored as node-level priors for SEPAL. For each center spot, we constructed a two-hop spatial ego-graph using six neighbors for hexagonal Visium geometry and eight neighbors for grid geometry. Node features were formed by concatenating PFM embeddings with sinusoidal 2D positional embeddings, followed by a preprocessing MLP, GCNConv layers with hidden dimensions $[512,256,128]$, and SAGPooling. The graph branch predicted a residual correction that was added to the LocalNet prediction.

\noindent \textbf{SGN}:
SGN \cite{Yan_Spatial_MICCAI2024} implements a zero-shot semantic-guided framework for spatial gene expression prediction. In the original setting, window embeddings are extracted using ResNet-50 and propagated through a 4-layer GraphSAGE network with hidden dimension 512, using both spatial-neighborhood and feature-neighborhood edges. Gene semantic embeddings are obtained from large-language-model-derived gene descriptions (Neural-Chat-v3 \cite{mukherjee2023orca}; 4096-dimension) and processed by a Transformer-based weight generator to produce gene-specific prediction weights and biases. Expression levels are predicted by applying these gene-specific parameters to the graph-refined window features, and the model is trained with a mean-squared-error regression loss. 
For a fair comparison, we replaced the ResNet-50 window embeddings with PFM embeddings, while retaining the GraphSAGE propagation and semantic gene-specific prediction mechanism.

\noindent \textbf{TRIPLEX}:
TRIPLEX \cite{chung2024accurate} integrates local, neighborhood, and global histological context for spatial gene expression prediction. It consists of three branches: a target branch that encodes the local H\&E patch image centered at each spot, a neighbor branch that processes image embeddings from surrounding spots, and a global branch that models slide-level image embeddings together with spatial positional information. The three representations are combined through a cross-attention-based fusion encoder, and gene expression is predicted from the fused representation. TRIPLEX also applies auxiliary supervision to the target, neighbor, and global branches, together with a consistency loss between the fused prediction and each branch-specific prediction. 
In the original implementation, all three branches rely on ResNet18 trained on histology images \cite{ciga2022self} as the image encoder. In our adaptation, we replaced the neighbor and global branch encoders with PFM-derived image embeddings, as TRIPLEX already accepts precomputed embeddings for these two branches. The target branch was kept as the original ResNet18 encoder, since it produces spatial visual tokens directly from the raw H\&E images. For the global branch position encoding, we adopted SpatialFormer \cite{xiao2024spatialformer}, supported in the official repository. The model was configured with an embedding dimension to match the PFM embedding dimensionality, with transformer depths of 1/2/1 for the fusion, global, and neighbor encoders respectively, 8 attention heads and MLP ratio of 2 in all branches, and a dropout rate of 0.15 in the fusion encoder.

\noindent \textbf{DeepSpot}:
DeepSpot \cite{nonchev2025deepspot} is a deep-set model composed of $\phi_{\mathrm{spot}}$ and $\rho_{\mathrm{gene}}$ modules. Each spot is represented using three types of histology-derived features: the target spot embedding, embeddings from a $3\times3$ grid of non-overlapping sub-spots, and embeddings from neighboring spots. A shared $\phi_{\mathrm{spot}}$ module transforms these feature sets, with sub-spot features aggregated by summation and neighbor features aggregated by max pooling. The resulting target, sub-spot, and neighbor representations are concatenated and passed to $\rho_{\mathrm{gene}}$ for multivariate gene expression regression. In the original study, DeepSpot was evaluated using PFM-derived features from models such as UNIv1, Phikon, and H-Optimus0. 
For consistency with the other benchmarked models, target, sub-spot, and neighbor embeddings were extracted using the same PFM feature-extraction protocol, with UNIv2 used as the default feature extractor. DeepSpot was configured to use the full spot--subspot--neighbor context, with $\phi_{\mathrm{spot}}$ projecting each input embedding to a 512-dimensional representation using dropout rate 0.3 and ReLU activation. We used an ensemble of 10 $\phi_{\mathrm{spot}}$ modules aggregated by median and 10 $\rho_{\mathrm{gene}}$ modules aggregated by mean.

\noindent \textbf{BLEEP}:
BLEEP \cite{xie2023spatially} employs a bi-modal contrastive learning framework that aligns histology image embeddings and gene expression profiles in a shared embedding space. In the original model, each H\&E image patch is encoded by an image backbone, such as an ImageNet-pretrained ResNet-50, and projected into the multimodal space through an MLP projection head. The corresponding gene expression vector is projected through a separate spot projection head, and the two modalities are aligned using a bidirectional contrastive loss with temperature set to 1.0. During inference, the cosine similarity between each query histology image embedding and all reference gene expression embeddings in the shared space is computed, and the top-$k$ reference spots with the highest similarity scores are selected. The gene expression vectors of these retrieved spots are then averaged to produce the final prediction.
For consistency with the other benchmarked models we replaced the image encoder with PFM-derived image embeddings. These PFM embeddings were passed through BLEEP's image projection head and contrastively aligned with the projected gene expression embeddings using the same training objective. The inference procedure was retained, using nearest-neighbor retrieval in the shared embedding space followed by averaging the expressions of the top-$k$ retrieved reference spots. In our experiments, we set $k = 50$ and used uniform (unweighted) averaging rather than similarity-weighted aggregation. Both image and spot embeddings are projected to a 512-dimensional shared space via a two-layer projection head with dropout of 0.1.

\noindent \textbf{STco}:
STco \cite{shi2024spatial} adopts the same contrastive retrieval framework as BLEEP, aligning histology image embeddings and gene expression profiles in a shared latent space via a bidirectional contrastive loss with temperature set to 1.0. STco augments each gene expression vector with learnable x- and y-coordinate embeddings before projection, incorporating spatial positional information into the expression encoder. During inference, each query image embedding retrieves the top-$K$ most similar reference spot embeddings in the learned space, and the final gene expression prediction is obtained by distance-weighted aggregation of the retrieved reference expressions. We replaced the image encoder with PFM-derived spot-level image embeddings while retaining the coordinate-aware gene expression branch, contrastive objective, and top-$K$ retrieval-based inference procedure. Spatial coordinates are encoded via learnable x- and y-position embeddings, with the vocabulary size set to exceed the maximum number of spots across all datasets. Both image and spot embeddings are projected to a 512-dimensional shared space via a two-layer projection head with dropout of 0.1.

\noindent \textbf{mclSTExp}:
mclSTExp \cite{min2024multimodal} is another contrastive retrieval model that extends the BLEEP framework. It augments each gene expression vector with learnable x- and y-coordinate embeddings and processes the result through a lightweight Transformer-based spot encoder before projection, enabling the model to capture spatially varying expression patterns beyond the coordinate-free alignment of BLEEP. During inference, gene expression is predicted by similarity-weighted averaging of the retrieved reference expressions. As in BLEEP and STco, the model is trained with a bidirectional contrastive loss between image and spot embeddings. To ensure a consistent comparison, we replaced the image encoder with a PFM, while retaining the coordinate-aware spot encoder, contrastive objective, and retrieval-based prediction procedure. We represented spatial position using independent learnable embeddings along the x- and y-axes, allocating sufficient embedding capacity to cover the full coordinate range across all datasets. The spot encoder consists of 2 Transformer layers with 8 attention heads, head dimension 64, and a dropout rate of 0.1. Both image and spot embeddings are projected to a 512-dimensional shared space, and contrastive alignment is performed with a temperature of 1.0.

\noindent \textbf{HistoSPACE}:
HistoSPACE \cite{kumar2024histospace} employs a lightweight convolutional autoencoder for unsupervised feature extraction from histology patches, followed by a linear decoder for spot-level expression prediction. The convolutional encoder consists of three blocks of 3$\times$3 convolution, batch normalization, ReLU activation, and max-pooling, progressively expanding channels from 3 to 32, 64, and 128. A symmetric transposed-convolution decoder reconstructs the input image from the bottleneck representation. For expression prediction, an additional convolutional layer and max-pooling are applied to the encoder output, followed by two fully connected layers (hidden dimension 256) that map to gene expression values. Whereas the original implementation adopts a two-stage procedure --- first pretraining the autoencoder with image reconstruction loss, then fine-tuning only the expression head with the encoder frozen --- we trained the full model end-to-end in a single stage using MSE loss over gene expression prediction alone, for simplicity and consistency with the unified training protocol applied across all benchmarked models.

\noindent \textbf{Path2Space}:
Path2Space \cite{shulman2025path2space} predicts spot-level gene expression via direct supervised regression. Each $224 \times 224$ image patch centered on an ST spot is encoded by a frozen pathology foundation model (CTransPath in the original work) into a fixed-dimensional feature vector, which is then mapped to gene expression values through an MLP regressor trained with mean squared error loss. This architecture is functionally equivalent to linear probing when a single fully connected layer is used, and serves as a straightforward baseline for evaluating the informativeness of the underlying image encoder. We replaced CTransPath with UNIv2 as the default PFM, consistent with the feature-extraction protocol applied to all other models. The MLP regressor comprises three fully connected layers with ReLU activations and a hidden dimension matching the PFM embedding dimensionality, with no dropout applied.

\noindent \textbf{M2ORT}:
M2ORT \cite{wang2024m2ort} predicts spatial gene expression by jointly exploiting multi-scale histology images corresponding to the same spatial spot. Images from different WSI magnification levels are embedded with level-dependent image embeddings to form aligned token sequences. Intra-Level Token Mixing Modules (ITMM) apply self-attention independently within each scale to capture scale-specific morphological features, while an Inter-Level Channel Mixing Module (ICMM) enables cross-scale feature fusion through channel-wise mixing. The final gene expression is predicted by concatenating the multi-scale [CLS] tokens and applying a linear regression head trained with mean squared error loss. As multi-scale image inputs are integral to this architecture, replacing the scale-dependent encoders with a single PFM is structurally incompatible; we therefore retained the original encoder design. The model was configured with three resolution streams, each with a hidden dimension of 256, combined into a shared dimension of 768, along with 8 Transformer layers, 12 attention heads, a head dimension of 64, and an MLP dimension of 512. CLS token pooling was applied for gene expression prediction.

\noindent \textbf{M2OST}:
M2OST \cite{wang2025m2ost} adopts a many-to-one regression Transformer for spot-level spatial gene expression prediction by jointly leveraging multi-resolution pathology images. For each target spot, image patches from multiple magnification levels are extracted with a fixed spatial size and encoded in parallel using a shared deformable patch embedding scheme that emphasizes the central spot region while preserving surrounding contextual information. Each resolution stream is processed independently by intra-level self-attention to capture scale-specific representations, followed by cross-level token and channel mixing modules that enable information exchange across resolutions. The final representations from all scales are concatenated and passed to a linear regression head to predict gene expression in an end-to-end supervised manner. As this architecture depends on simultaneous multi-resolution image streams with a shared deformable patch embedding, substituting the encoder with a single PFM is structurally incompatible; we therefore retained the original encoder design. The architecture was configured with a hidden dimension of 768 across three resolution streams (256 per stream), 8 transformer layers, 12 attention heads, head dimension 64, and MLP dimension 512, with CLS token pooling and a batch size of 32.

\noindent \textbf{Stem}:
Stem \cite{zhu2025diffusion} formulates spatial gene expression prediction as a conditional generative modeling problem rather than deterministic regression. For each ST spot, histology features extracted from a PFM are used as conditioning signals. Gene expression is represented as a sequence of gene tokens, where each token combines a learnable gene identity embedding with a count-value embedding. A Diffusion Transformer (DiT) backbone is trained under a denoising diffusion framework to model the conditional distribution of gene expression given the histology condition. The histology condition is injected into the DiT blocks through adaptive LayerNorm modulation together with the diffusion timestep embedding, and gene expression is generated by reverse diffusion at inference time. In the original implementation, the histology condition is formed by concatenating embeddings from two PFMs, UNIv1 and CONCH \cite{lu2024visual}, and image augmentation is applied before feature extraction. For a fair comparison with the other benchmarked models, we replaced this dual-encoder conditioning with a single UNIv2 embedding and omitted augmentation during feature extraction, consistent with the feature-extraction protocol applied to all other models. The DiT backbone was configured with a hidden dimension of 768, 4 transformer layers, 16 attention heads, and an MLP ratio of 4.0. The diffusion process uses 1000 timesteps with a linear noise schedule, predicts the noise (\texttt{$\epsilon$-prediction}), and learns the variance (\texttt{LEARNED\_RANGE}), trained with a hybrid MSE and KL loss. An exponential moving average (EMA) of the model weights is maintained and used at inference time. The model was trained with a batch size of 8.

\noindent \textbf{STFlow}:
STFlow \cite{huang2025scalable} formulates spatial gene expression prediction as a flow-matching generative task. It learns to iteratively transform samples from an expression prior into spatial gene expression profiles conditioned on histology image-derived embeddings and spatial coordinates. In the original model, histology image embeddings are extracted using a PFM and used as conditioning inputs. STFlow adopts a zero-inflated negative binomial (ZINB) prior to reflect the sparsity and overdispersion of spatial transcriptomics data. During training, the model samples interpolation states between prior samples and ground-truth expression, and a denoising network predicts the target expression from the interpolated state, image embeddings, coordinates, and timestep using a mean-squared-error objective. The denoiser is implemented as a frame-averaging spatial Transformer that aggregates local information from neighboring spots through E(2)-invariant attention. 
Since STFlow originally conditions the flow-matching model on PFM-derived histology features, we did not modify the core architecture. Instead, we standardized the conditioning features by varying the PFM under the same benchmarking protocol, with UNIv2 used as the default feature extractor. For all experiments, the denoiser was configured as a 4-layer spatial Transformer with 4 attention heads, a hidden dimension of 128, and a pairwise hidden dimension of 128, using SwiGLU activations and 8 nearest neighbors for local E(2)-invariant attention. The ZINB prior was parameterized with total count of 1, logits of 0.1, and zero-inflation logits of 0, with prior samples log1p-normalized before interpolation. At inference, gene expression is generated using 5 Euler integration steps.

\noindent \textbf{OmiCLIP}:
OmiCLIP \cite{chen2025visual} formulates histology–transcriptomics integration as a multi-modal contrastive learning problem. For each Visium spot, the paired H\&E image patch is encoded by a pathology image encoder, while the corresponding transcriptomic profile is converted into a gene “sentence” by ranking highly expressed genes and concatenating their gene symbols. Image and transcriptomic encoders are jointly trained using a CLIP-style contrastive objective to align paired image–gene embeddings in a shared latent space, with rank-based gene representations to mitigate batch effects and sequencing depth variability. The learned visual–omics embedding space serves as a foundation for downstream spatial analysis tasks. As OmiCLIP is pretrained on large-scale multi-institutional data as a vision--omics foundation model, we evaluated it in a zero-shot setting without any task-specific fine-tuning on our benchmark datasets. Specifically, the ViT-L/14 backbone pretrained by CoCa \cite{yu2022coca} strategy is used to encode spot-level H\&E patch images and gene sentences, where each gene sentence is constructed from the top-50 expressed genes per spot after excluding housekeeping genes. Gene expression is then predicted for each test spot via similarity-weighted averaging: the dot-product similarity between the test image embedding and all training spot gene-sentence embeddings is computed, and the predicted expression is obtained as the similarity-weighted mean of the corresponding training spot expressions.

\noindent \textbf{STPath}: 
STPath \cite{huang2025stpath} formulates spatial transcriptomics prediction as a generative masked modeling task. Each spatial spot is represented by aggregating histology features from a pathology foundation model (GigaPath\cite{xu2024whole}), gene expression, organ type, and sequencing technology embeddings into a unified token. Spot tokens are processed by a geometry-aware Transformer with E(2)-invariant spatial self-attention based on frame averaging (4 layers, 4 heads, hidden size = 512). Training masks gene expression values using a Beta-scheduled masking strategy and optimizes mean-squared-error loss to recover masked expressions. As STPath is pretrained at scale across diverse tissues and sequencing platforms as a ST foundation model, we evaluated it in a zero-shot setting without any task-specific fine-tuning on our benchmark datasets.

\heading{Training and evaluation details}

\noindent All models requiring task-specific training were implemented in PyTorch 2.3.1 and trained on a single NVIDIA RTX A5000 GPU.
The learning rate was set to 1 × 10$^{-4}$ with a maximum of 200 epochs.
Early stopping was applied with patience = 20, monitoring Pearson correlation. The epoch demonstrating the highest Pearson correlation coefficient was selected as the optimal epoch.
Learning-rate scheduling followed \texttt{ReduceLROnPlateau} (regression; patience = 5, factor = 0.1) or \texttt{CosineAnnealingLR} (contrastive/generative; $\eta\_min$ = 10$^{-6}$). 

\noindent To ensure a fair and consistent comparison, we adopted a unified cross-validation strategy across all 21 models, rather than following the validation scheme originally proposed in each method. Fold splits were constructed at the patient level to prevent data leakage from slides of the same patient appearing in both training and validation sets. For the HEST-bench datasets, we retained the original train/validation splits provided by the benchmark. For all remaining datasets, we applied $k$-fold cross-validation with $k=5$. Within each fold, models were trained on the training partition ($k-1$ folds) and evaluated on the held-out validation partition ($1$ fold); the final internal performance was obtained by averaging across all folds.

\noindent We define two categories of model architectures according to their training granularity: slide-based and patch-based models. Slide-based models predict gene expression for all patches within a single whole-slide image (WSI) and compute the loss over the entire slide. In contrast, patch-based models sample patches across multiple slides to form a mini-batch, predict gene expression for the patches within each batch, and compute loss independently for those patches. For slide-based models (EGGN, SGN, STFlow, HisToGene), the batch size was fixed at 1 to process complete slides as individual training samples, while patch-based models were trained with a larger batch size of 128, unless otherwise specified. For multimodal models (BLEEP, STco, mclSTExp), contrastive temperatures were fixed at 1.0.
During the testing phase, an evaluation was conducted on a per-slide basis for both the patch-based model and the slide-based model. The mean performance was computed across the samples within each fold. Subsequently, the results were averaged over all folds to derive the final model performance metric.

\noindent All histology image patches corresponding to spatial transcriptomic spots were processed using a standardized augmentation and normalization pipeline implemented with TorchVision.transforms. During training, each patch was first converted from a \texttt{NumPy} array to a \texttt{PIL} image, then subjected to a series of stochastic geometric augmentations: random horizontal and vertical flips (probability = 0.5 each), followed by a random 90° rotation applied with 50\% probability.

\noindent Following augmentation, images were converted to tensors normalized using ImageNet statistics (mean = [0.485, 0.456, 0.406]; standard deviation = [0.229, 0.224, 0.225]). This normalization was consistently applied to all models that receive image-based inputs. For validation and inference, only deterministic transformations (image normalization) were applied to preserve the original tissue morphology. 

\noindent All other implementation details not explicitly described above were faithfully reproduced following the original publications of each model to ensure consistency with reported architectures and training strategies.

\heading{Performance metrics}

\noindent \textbf{Pearson correlation coefficient (PCC).}
Spot-level prediction accuracy was quantified using the Pearson correlation coefficient, computed independently for each gene across all capture spots within a section. For gene $g$ in section $s$ with $N$ spots:
\begin{equation}
\text{PCC}_{g,s} = \frac{\sum_{i=1}^{N}(\hat{y}_{i,g} - \bar{\hat{y}}_{g})(y_{i,g} - \bar{y}_{g})}{\sqrt{\sum_{i=1}^{N}(\hat{y}_{i,g} - \bar{\hat{y}}_{g})^2 \cdot \sum_{i=1}^{N}(y_{i,g} - \bar{y}_{g})^2}},
\end{equation}
where $\hat{y}_{i,g}$ and $y_{i,g}$ denote the predicted and ground-truth log-normalized expression of gene $g$ at spot $i$, respectively. PCC was computed using the \texttt{PearsonCorrCoef} implementation in TorchMetrics (v1.9.0) with \texttt{num\_outputs} set to the number of evaluated genes, enabling efficient GPU-accelerated computation across all genes simultaneously. NaN values arising from genes with zero variance (constant expression) were set to zero prior to aggregation. Section-level PCC was obtained by averaging gene-wise values, and dataset-level PCC was reported as the mean across all held-out sections within each cross-validation fold, further averaged across folds. Standard deviation was calculated across cross-validation folds.
 
\noindent \textbf{Mean absolute error (MAE).}
To quantify the magnitude of prediction error in expression space, MAE was computed per gene across all spots within each section:
\begin{equation}
\text{MAE}_{g,s} = \frac{1}{N} \sum_{i=1}^{N} \left| \hat{y}_{i,g} - y_{i,g} \right|.
\end{equation}
MAE was computed using the \texttt{MeanAbsoluteError} implementation in TorchMetrics with per-gene outputs. Aggregation followed the same scheme as PCC: gene-wise values were averaged to yield a section-level score, and dataset-level MAE was summarized as the mean and standard deviation across sections and folds. As MAE operates in log-normalized expression space, its values reflect absolute deviations in $\log(1+x)$-transformed counts.
 
\noindent \textbf{Spatial structural similarity index (SSIM).}
To capture the spatial coherence of predicted expression landscapes beyond spot-level accuracy, we additionally computed SSIM, which jointly accounts for local luminance, contrast, and structural patterns. For each tissue section and each gene, predicted and ground-truth expression values were projected onto a two-dimensional spatial grid. Spot coordinates were rescaled to discrete grid indices via a rescaling factor $r = d_{\text{center}} / s_{\text{pixel}}$, where $d_{\text{center}}$ is the nominal center-to-center distance between adjacent capture spots ($100\,\mu$m for 10x Visium) and $s_{\text{pixel}}$ is the section-specific pixel size ($\mu$m/pixel) retrieved from a precomputed metadata table. Rescaled coordinates were zero-indexed to obtain integer grid positions $(x', y')$, and expression values were placed into zero-initialized matrices of dimensions $(\max(x') + 1) \times (\max(y') + 1)$. Both ground-truth and predicted expression images were independently min--max normalized to the $[0, 1]$ range. SSIM was then computed using \texttt{skimage.metrics.structural\_similarity} with \texttt{data\_range=1.0} and all other parameters set to their default values. Gene-level SSIM values were averaged across genes, excluding NaN entries from degenerate constant images, to yield a section-level score. These section-level scores were then averaged across all samples within each fold. Dataset-level SSIM was reported as the mean and standard deviation across all cross-validation folds.
 
\noindent \textbf{Summary statistics.}
All three metrics were computed on GPU (NVIDIA) where applicable. For benchmark comparisons, mean values across cross-validation folds were used as the primary performance estimate, with standard deviation reported to reflect inter-fold variability. Higher PCC and SSIM values indicate better predictive performance, whereas lower MAE indicates smaller absolute prediction error.

\heading{Benchmarking method details}

\noindent To train and evaluate the models, we used two distinct gene sets: (1) high-mean highly-variable genes (HMHVG), and (2) tumor microenvironment (TME) marker genes.

\noindent \textbf{HMHVG selection}: To select the gene set used for performance evaluation, we adopted the gene selection strategy from Zhu et al. \cite{zhu2025diffusion}. First, we identified genes shared across all ST slides by taking the intersection of their gene symbols. Each slide was processed independently: we filtered out spots with zero detected genes and genes expressed in zero spots, followed by total-count normalization and log-transformation. Using these processed expression profiles, 2,000 highly variable genes (HVGs) were computed per slide via the Scanpy function \texttt{sc.pp.highly\_variable\_genes}. We then constructed a candidate gene pool by taking the union of HVGs across all slides, subsequently excluding mitochondrial and ribosomal genes based on their name prefixes (MT/mt-, RPS-, and RPL-). To derive the final HMHVG set, we aggregated the raw counts of the candidate genes across all spots and slides. For each candidate gene, we computed the mean expression and standard deviation across all spots and slides, and ranked genes in descending order for each metric independently. A rank-sum score was then derived by summing the two rank values, and the 200 genes with the lowest rank-sum scores were selected as the final HMHVG set, representing genes that are simultaneously highly expressed and highly variable across the dataset.

\noindent \textbf{TME marker gene selection}: To define the TME-focused marker panel, we started from the 10x Genomics Human Immune Cell Profiling Panel antibody list\cite{10x_panel}, which provides 31 human target markers annotated by cell type or marker category, excluding isotype controls. We then intersected this curated marker list with the genes available in our spatial transcriptomics datasets and retained the overlapping genes for evaluation. This procedure yielded a final set of 16 marker genes spanning distinct functional lineages, including epithelial markers (\textit{KRT5}); stromal and vascular markers (\textit{ACTA2}, \textit{VIM} and \textit{PECAM1}); T-lymphocyte markers (\textit{CD3E}, \textit{CD4}, \textit{CD8A}, and \textit{CXCR5}); B-lymphocyte markers (\textit{CD19} and \textit{CCR7}); myeloid lineage markers (\textit{CD68}, \textit{CD163}, \textit{ITGAX}, and \textit{ITGAM}); and immune checkpoint markers (\textit{PDCD1} and \textit{CD274}).

\noindent \textbf{Patch encoder utilization}: To assess the impact of the patch-level image encoder on prediction performance, we evaluated five representative encoders spanning a range of pretraining strategies and architectures: ResNet-50 \cite{he2016deep} (pretrained on ImageNet dataset \cite{deng2009imagenet}), CTransPath \cite{wang2022transformer}, UNIv2 \cite{chen2024towards}, Virchow2 \cite{zimmermann2024virchow}, and H-Optimus1 \cite{hoptimus1}. During preprocessing, image patches were normalized strictly according to the specific normalization protocols originally established for each respective encoder. Patch-level features for all models were then extracted offline using the \texttt{Trident} \cite{zhang2025accelerating} tool, which provides a unified and efficient interface for diverse pathology encoders. Notably, to maintain comparable feature dimensionality across models and prevent excessive computational overhead, we modified the feature extraction strategy for Virchow2; instead of utilizing its default representation (concatenation of the \texttt{CLS} token and mean pooling), we exclusively extracted the \texttt{CLS} token. The extracted features from all models were systematically stored as \texttt{HDF5} files, enabling efficient reuse across model evaluations. Throughout these comparative analyses, all common training hyperparameters were held constant.

\noindent \textbf{Internal test}: For internal evaluation, each model was trained and assessed using $k$-fold cross-validation within each of the seven training datasets. Specifically, training data were partitioned into $k$ folds such that each sample appeared in the test split exactly once; the fold assignments were fixed across all models to ensure comparability. Gene expression predictions were generated for each held-out sample, and per-gene performance metrics between predicted and observed expression were computed. The primary performance metric was the mean Pearson correlation coefficient averaged across genes and folds, supplemented by mean absolute error (MAE) and structural similarity index (SSIM). All experiments were run with a fixed random seed (2021) for reproducibility. To assess performance across expression levels, predicted genes were stratified into five quantiles based on mean expression, and the Pearson correlation was computed within each quantile. This analysis was conducted for all models to determine whether prediction accuracy varies systematically with expression level and whether different model architectures are differentially effective for low- versus high-expression genes. 

\noindent \textbf{External test}: To evaluate generalization, we assessed models trained on each internal dataset against seven external test datasets that were not used during training. External validation scenarios were grouped into three categories: (i) \textit{cross-institution}, where the training and test tissues are from the same tumor type but different institutions; (ii) \textit{cross-tissue}, where the model is applied to a different tumor type than that seen during training; and (iii) \textit{cross-platform}, where the test data were acquired on a different ST platform than that used for training (e.g., models trained on Visium data evaluated on Xenium data). For inference, the model checkpoint from fold 0 was used. Predictions were computed over the common gene set between the training and test datasets. Per-gene Pearson correlations were then computed and averaged over the shared genes. We further quantified the generalization gap as the difference between external and internal mean Pearson correlation, and examined its correlation with internal performance across models using Spearman correlation analysis to assess whether internal rankings are predictive of external performance.

\noindent \textbf{Gene-level evaluation}: To assess gene-level prediction accuracy, we performed the analysis on the evaluated gene panel for each dataset, defined as the union of HMHVGs and marker genes. For each model--dataset pair, we computed the per-gene Pearson correlation coefficient (PCC) between predicted and ground-truth expression values across all spots in each section, averaged these values across all sections within each cross-validation fold, and then averaged across all folds. To identify genes for which models demonstrated substantially superior prediction performance, we applied a margin-based selection procedure using the linear probing baseline as the reference. For each gene, we identified the non-baseline model achieving the highest margin over the baseline. A gene was retained if: (i) its baseline PCC was finite; (ii) the best competing model achieved a PCC $\geq$ 0.4; and (iii) the margin exceeded 0.1. Genes were ranked by margin, and the top eight genes were selected for Fig. \ref{fig:gene_analysis}b. Selected genes were manually assigned to one of four functional categories---\textit{Stromal/ECM}, \textit{Immune/Inflammation}, \textit{Epithelial}, and \textit{Stress/Proliferation}---based on established marker gene lists from the literature. Genes not matching any category were labeled \textit{Other}.

\noindent \textbf{Single-sample gene set scoring}: To evaluate whether virtual ST predictions preserve gene set level biological activity beyond individual gene level accuracy, we performed gene set level prediction analysis using the evaluated gene panel. Specifically, we used the union of HMHVGs and marker genes as the available gene set for each dataset. Gene sets were obtained from the Gene Ontology Biological Process library, \texttt{GO\_Biological\_Process\_2023}\cite{gene2023gene}. 
For each GO Biological Process term, we retained only gene set with sufficient overlap with the available genes. A gene set was included if at least five of its member genes were present in the available gene set and if the overlapping genes covered at least 10\% of the original gene set gene members. This filtering ensured that gene set scores were computed only for gene sets with adequate representation in the measured and predicted gene space. For each sample, both the ground-truth ST expression and the predicted virtual ST expression matrices were converted from spot-by-gene matrices into spot-by-gene set activity matrices. Gene set scores were estimated using singscore \cite{foroutan2018single}.
We then compared gene set score profiles between the ground-truth ST and virtual ST predictions. For each gene set, Pearson correlation was computed across spatial spots between the ground-truth gene set score vector and the predicted gene set score vector. The resulting gene set-level correlations were averaged across samples and folds for each model and dataset, providing a gene set level measure of how well each model preserved spatial biological programs in the predicted virtual ST profiles.

\noindent \textbf{Cell type deconvolution}: We evaluated whether virtual ST profiles preserve the spatial organization of cell-type composition by applying the cell deconvolution tool Cell2location \cite{kleshchevnikov2022cell2location} to both true and model-predicted spatial expression matrices.
For each dataset, we applied Cell2location separately to the ground-truth ST data and to each model-generated virtual ST dataset. In all runs within a dataset, matched single-cell RNA-seq reference was used.
For LUAD and BRCA datasets, we used the single-cell Lung Cancer Atlas (LuCA) \cite{salcher2022high} and Human Breast Cell Atlas (HBCA) \cite{reed2024single} as reference single-cell RNA-seq datasets, respectively.
Gene sets were restricted to HMHVGs and marker genes. For both the ground-truth and virtual ST results, Pearson correlation was calculated for each cell type.
The Cell2location model was first trained on the reference single-cell RNA-seq datasets to estimate cell-type expression signatures, after which spatial cell abundance was inferred independently for each true or virtual ST sample. For virtual ST inputs, predicted expression matrices were inverse-transformed using expm1 to recover count-scale values, which were subsequently rounded and clipped to non-negative integers prior to inference.
The reference regression model was trained for 250 epochs with a batch size of 2,500, and posterior cell-type expression signatures were estimated from 1,000 posterior samples. For spatial mapping, Cell2location was run with \texttt{N\_cells\_per\_location} = 30, \texttt{detection\_alpha} = 200, \texttt{batch\_size} of 2,048, for 30,000 epochs using 1,000 posterior samples.
To mitigate the impact of limitations inherent to deconvolution-based estimation, samples were excluded from the final aggregate analysis when the highest mean Pearson correlation achieved by any virtual ST model across cell types was below 0.1.

\noindent \textbf{Spatial domain identification}: To evaluate whether virtual ST predictions preserve spatially organized tissue architecture, we applied SpaGCN \cite{hu2021spagcn} to both ground-truth and model-predicted spatial expression matrices to identify spatial domains. 
SpaGCN was run without histology image features, relying solely on spatial coordinates and gene expression. Gene sets were restricted to HMHVGs and marker genes, consistent with the evaluated gene panel. For ground-truth data, the optimal number of spatial domains was selected from candidates $k \in {3, 5, 7, 9}$ based on the silhouette score computed on the SpaGCN embedding. Domain boundaries were subsequently refined using square-grid neighbor refinement. For virtual ST predictions, the same number of clusters determined from the ground-truth run was applied to ensure a fair comparison, and predicted expression matrices were inverse-transformed with expm1 and clipped to non-negative integers prior to clustering.              To quantify the agreement between ground-truth and predicted spatial domain assignments, we computed the Adjusted Rand Index (ARI) between the ground-truth domain labels and those inferred from each virtual ST prediction. These metrics were averaged across samples and folds for each model and dataset.


\clearpage



\section*{Data availability} 

All datasets used in $\oursint$ and $\oursext$, with the exception of MGB-PRAD, are publicly available through their respective sources. The NCCHE lung adenocarcinoma dataset (Visium and Xenium) is available at \url{https://kero.hgc.jp/Ad-SpatialAnalysis_2024.html}. The WUSTL dataset (breast cancer, pancreatic ductal adenocarcinoma, and renal cell carcinoma) is available through the Human Tumor Atlas Network at \url{https://data.humantumoratlas.org/publications/hta12_2024_nature_chia-kuei-mo}. The Massey triple-negative breast cancer dataset is available via NCBI GEO under accession number GSE210616 (\url{https://www.ncbi.nlm.nih.gov/geo/query/acc.cgi?acc=GSE210616}). The SNU glioblastoma dataset is available at \url{https://zenodo.org/records/17572905}. The HEST-1K datasets used in this study are available via Hugging Face at \url{https://huggingface.co/datasets/MahmoodLab/hest}. The MGB-PRAD dataset was collected at Mass General Brigham under institutional review board approval and is not publicly available due to patient privacy restrictions; researchers interested in accessing this dataset may contact the corresponding author to inquire about data sharing agreements. Preprocessed versions of all publicly available datasets are deposited on Hugging Face at \url{https://huggingface.co/datasets/nexgem/STP-Bench}.

\section*{Code availability} 

The unified implementation of the 21 benchmarked models, and a modular pipeline for reproducing the full $\ours$ evaluation workflow are released at \url{https://github.com/NEXGEM/STP-Bench}.


\section*{Acknowledgements}
This work was supported in part by the Korea-US Collaborative Research Fund (KUCRF) grant funded by the Ministry of Science and ICT and the Ministry of Health \& Welfare, Republic of Korea (No. RS-2024-00468417); the Institute of Information \& Communications Technology Planning \& Evaluation (IITP) grant funded by the Korean government (MSIT) (No. RS-2019-II190421, AI Graduate School Support Program at Sungkyunkwan University); the Korea Health Technology R\&D Project through the Korea Health Industry Development Institute (KHIDI), funded by the Ministry of Health \& Welfare, Republic of Korea (No. RS-2025-02309552); and the National Research Foundation of Korea (NRF) grant funded by the Ministry of Education (MOE) of the Republic of Korea (No. RS-2025-25427169). Additional support was provided by the Brigham and Women’s Hospital (BWH) President’s Fund, Mass General Hospital (MGH) Pathology, and the National Institutes of Health (NIH) National Institute of General Medical Sciences (NIGMS) under award number R35GM138216.
\clearpage
\newcounter{extendeddatafigure}
\renewcommand{\theextendeddatafigure}{\arabic{extendeddatafigure}}

\begin{figure*}
\centering
\includegraphics[width=0.97\textwidth]{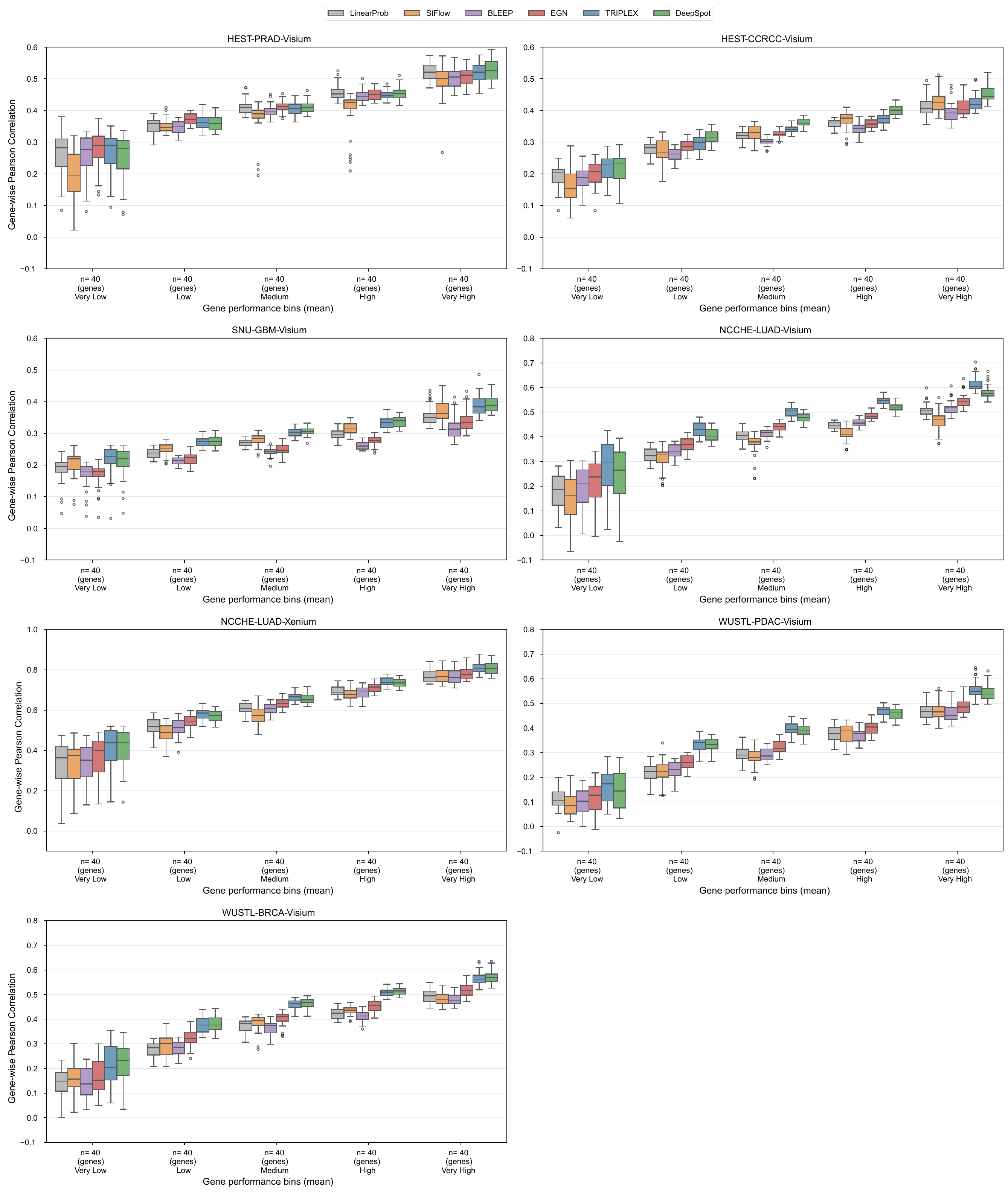}
\refstepcounter{extendeddatafigure}
\caption*{\textbf{Extended Data Fig.~\theextendeddatafigure: Gene-wise spatial expression prediction performance stratified by predictability for HMHVG}. Gene-wise Pearson correlation coefficient (PCC) distributions of five ST prediction models (STFlow, BLEEP, EGN, TRIPLEX, and DeepSpot) across gene predictability bins. 200 HMHVG genes per dataset are ranked by their mean PCC across all
   models and partitioned into five equal-frequency bins (Very Low to Very High). Each box shows the interquartile range of per-gene PCC for a given model within each bin; outliers are omitted for clarity.}
\label{fig:Genewise_hmhvg}
\end{figure*}

\begin{figure*}
\centering
\includegraphics[width=0.97\textwidth]{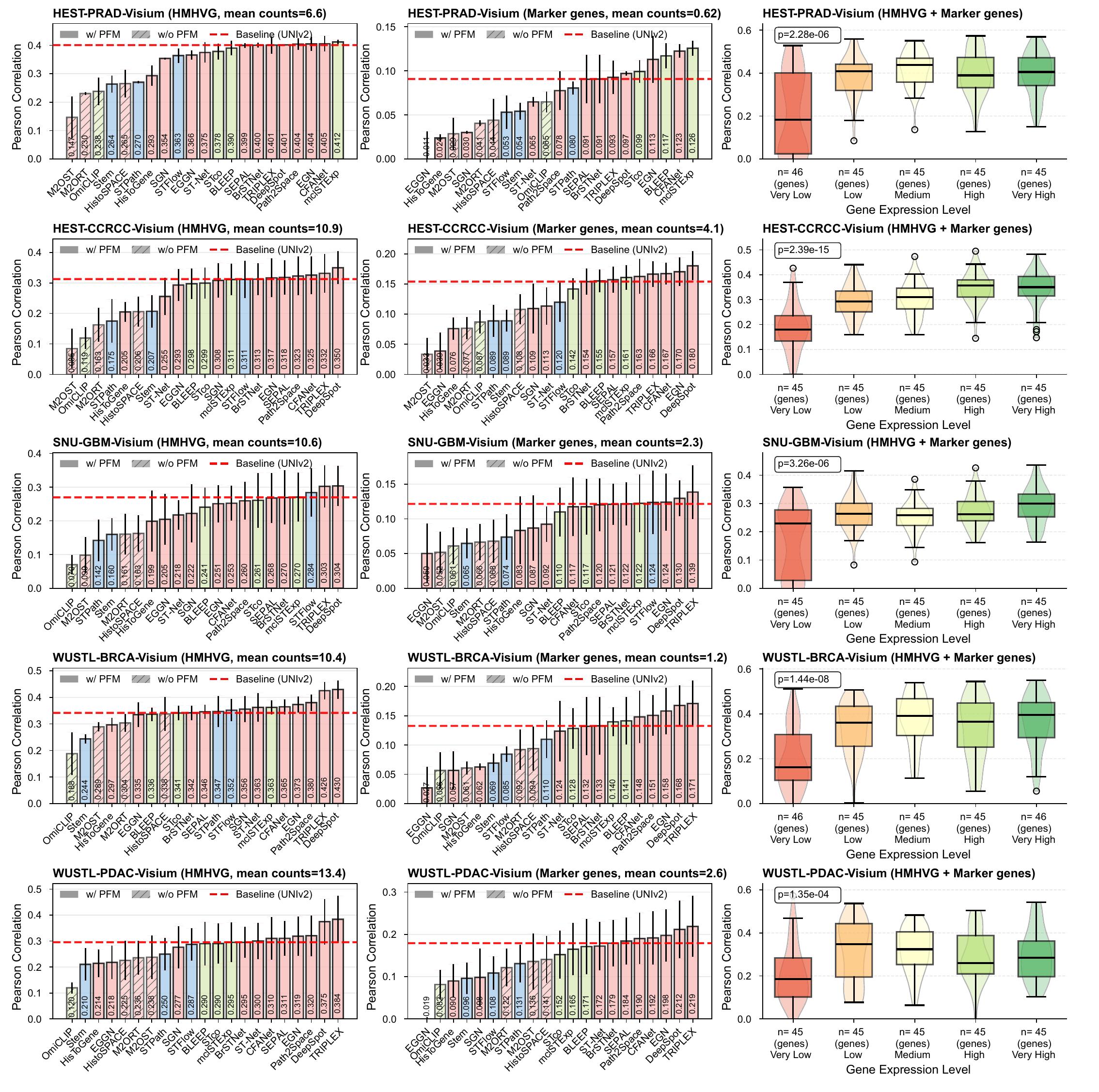}
\refstepcounter{extendeddatafigure}
\caption*{\textbf{Extended Data Fig.~\theextendeddatafigure: Dataset-wise performance of virtual ST models on $\oursint$.} 
Evaluation results corresponding to Fig. 2c–d, extended to all remaining datasets in STP-BENCH-INTERNAL. For each dataset, averaged gene counts per gene are shown. The bar charts (left) rank the models based on their Pearson correlation coefficient (PCC) performance for each dataset, with error bars indicating the standard deviation across cross-validation folds. The box plots (right) illustrate the distribution of prediction accuracy (PCC) across five gene expression levels (Very Low to Very High), with statistical significance assessed using the Kruskal–Wallis test.}
\label{fig:Data_wise_performance}
\end{figure*}

\begin{figure*}
\centering
\includegraphics[width=0.97\textwidth]{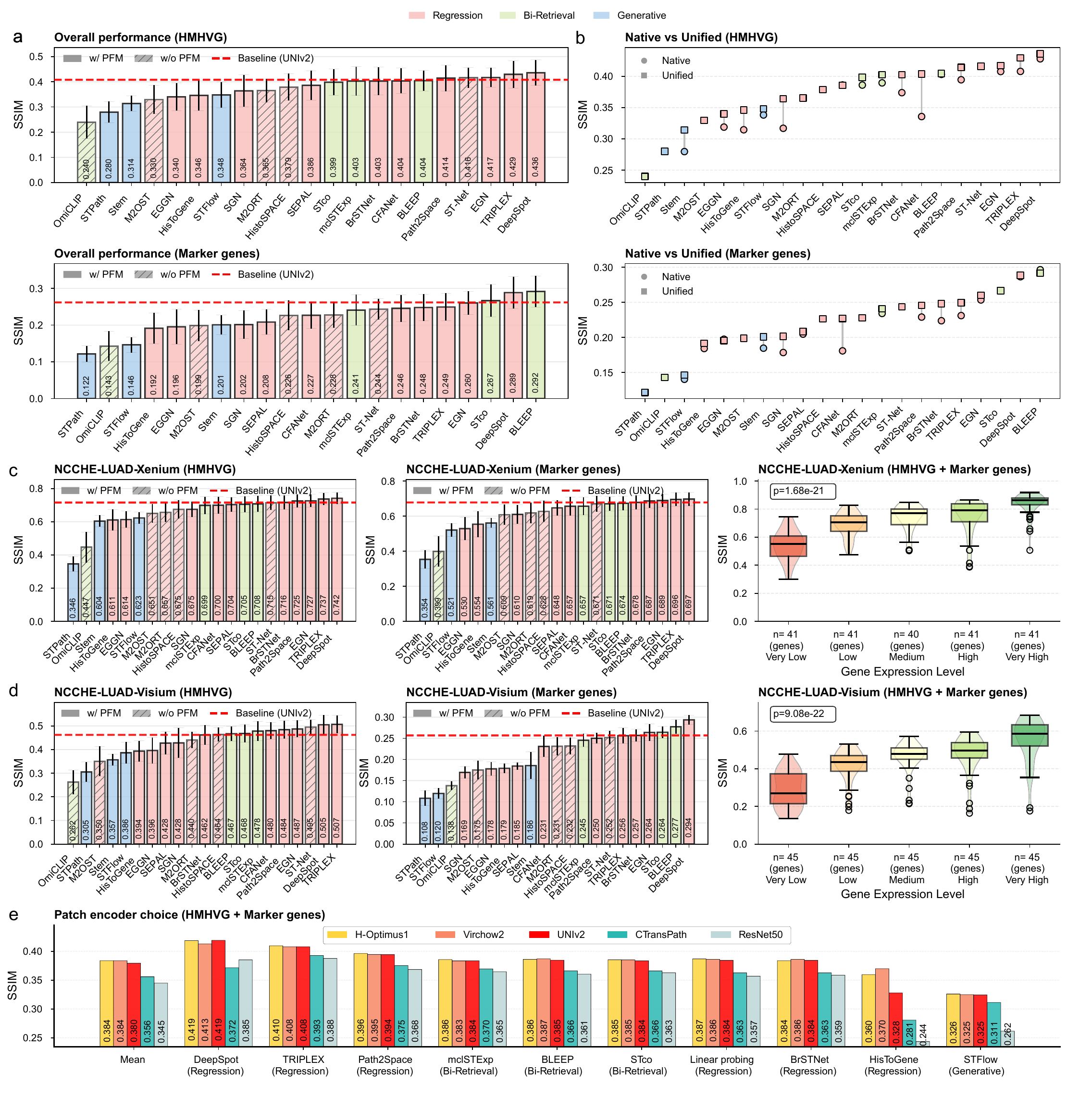}
\refstepcounter{extendeddatafigure}
\caption*{\textbf{Extended Data Fig.~\theextendeddatafigure: Performance benchmark of virtual ST baselines on $\oursint$ using SSIM}. 
\textbf{(a)} The average Structural Similarity Index Measure (SSIM) of virtual ST models evaluated on high-mean highly-variable genes (HMHVG) (top) and 16 marker genes (bottom), across 7 evaluation datasets. For each model, the patch features were extracted with UNIv2 (w/ PFM) unless a model required a specific patch feature extractor (w/o PFM, diagonal lines). The baseline (dashed line) represents the linear probing performance with UNIv2 image features. Error bars indicate the standard deviation across cross-validation folds.
\textbf{(b)} The comparison between virtual ST models when using UNIv2 feature extractor and the native feature extractors from respective original studies, evaluated by SSIM.
\textbf{(c-d)} Evaluation on representative NCCHE-LUAD-Xenium (c) and NCCHE-LUAD-Visium (d) datasets using SSIM. Averaged gene counts are shown for each gene across datasets. The bar charts (left) rank the models based on their SSIM performance on each specific dataset, with error bars indicating the standard deviation across cross-validation folds. The box plots (right) illustrate the distribution of prediction accuracy (SSIM) across five gene expression levels (Very Low to Very High), with Kruskal-Wallis test results indicating statistical significance.
\textbf{(e)} SSIM for ten selected models that utilize different patch encoders on the union set of HMHVG and marker genes: H-Optimus1, Virchow2, UNIv2, CTransPath, and ResNet50.
}
\label{fig:performance_SSIM}
\end{figure*}

\begin{figure*}
\centering
\includegraphics[width=0.8\textwidth]{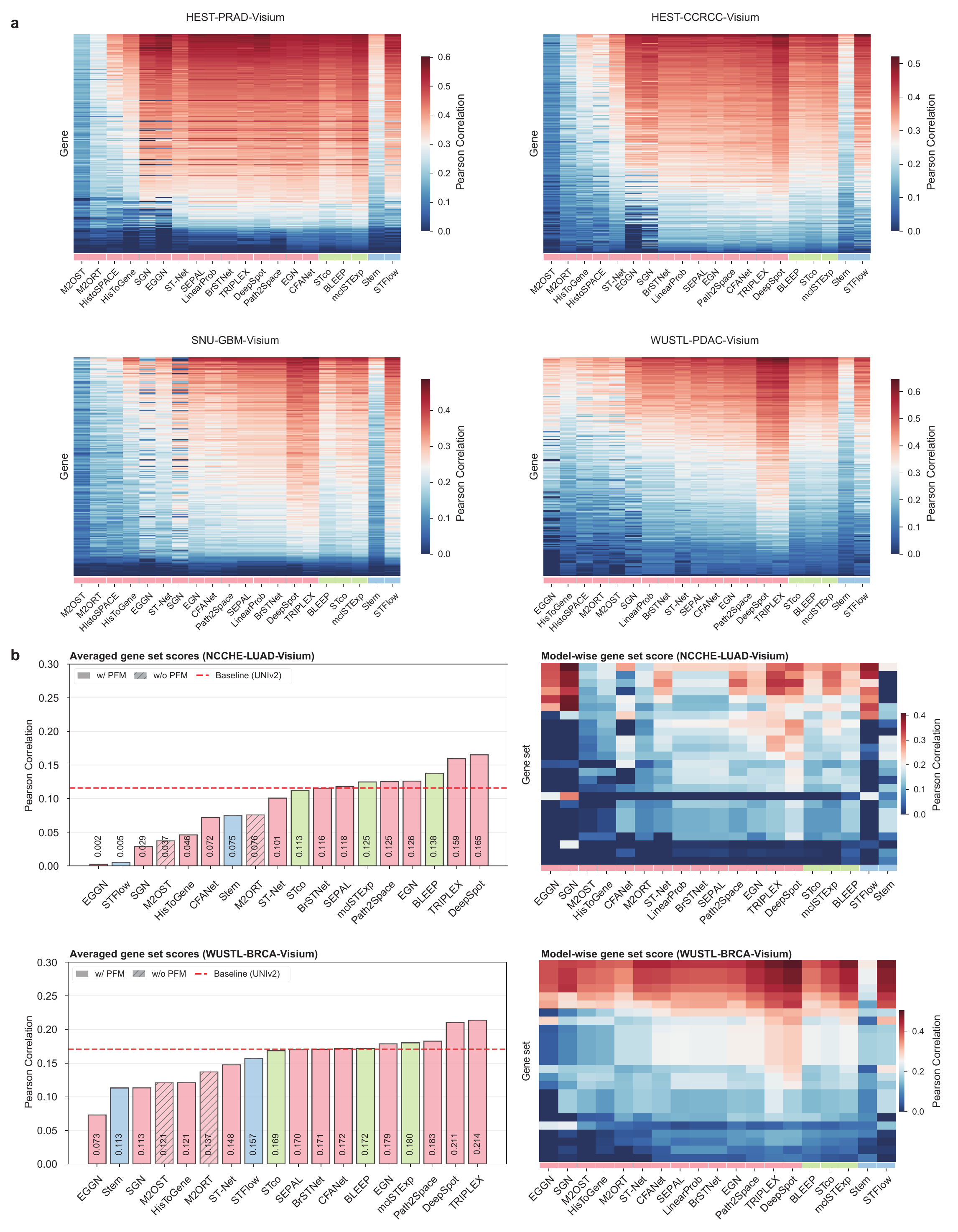}
\refstepcounter{extendeddatafigure}
\caption*{\textbf{Extended Data Fig.~\theextendeddatafigure: Gene- and gene-set-level prediction performance across additional datasets}. 
\textbf{(a)} Gene-wise performance heatmaps across benchmarked methods for additional datasets (HEST-PRAD, HEST-CCRCC, SNU-GBM, and WUSTL-PDAC). Each row and column represent individual genes (union of HMHVGs and marker genes) and prediction models, respectively, with the color bar indicating the methodological category of each model. The color scale denotes the gene-wise Pearson correlation coefficient (PCC).
\textbf{(b)} Gene-set Pearson evaluation in NCCHE-LUAD-Visium and WUSTL-BRCA-Visium. 
For each cohort, bar plots show model-wise average gene-set Pearson correlations, and heatmaps show per-gene-set PCCs across all benchmarked methods.
}
\label{fig:gene_geneset_add}
\end{figure*}

\begin{figure*}
\centering
\includegraphics[height=0.82\textheight,keepaspectratio]{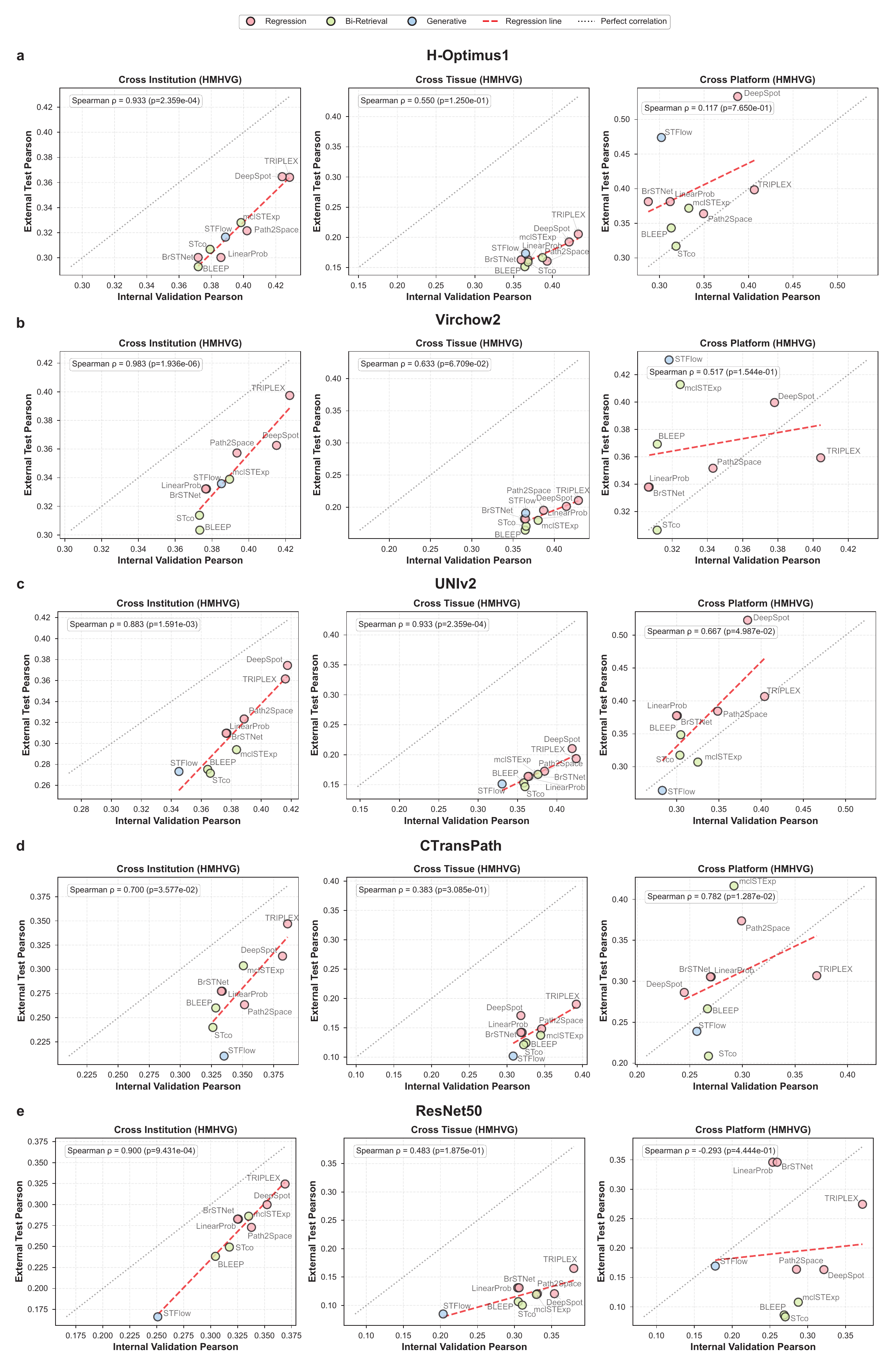}
\refstepcounter{extendeddatafigure}
\caption*{\textbf{Extended Data Fig.~\theextendeddatafigure: Impact of pathology foundation model choice on internal-to-external performance correlation of ST prediction models}. \textbf{(a-e)} Scatter plots illustrating the correlation between performance on $\oursint$ (x-axis) and $\oursext$ (y-axis) under three generalization scenarios (Cross-Institution, Cross-Tissue, and Cross-Platform), evaluated separately for five patch encoders: H-Optimus1, Virchow2, UNIv2, CTransPath, and ResNet50. Only models compatible with interchangeable patch encoders are included. The Spearman correlation coefficient was used to assess consistency of model rankings. Models are color-coded by category (Regression, Bi-Retrieval, Generative). The black dotted line indicates equal internal and external performance; the red line indicates the fitted linear regression.}
\label{fig:performance_corr}
\end{figure*}

\begin{figure*}
\centering
\includegraphics[height=0.82\textheight,keepaspectratio]{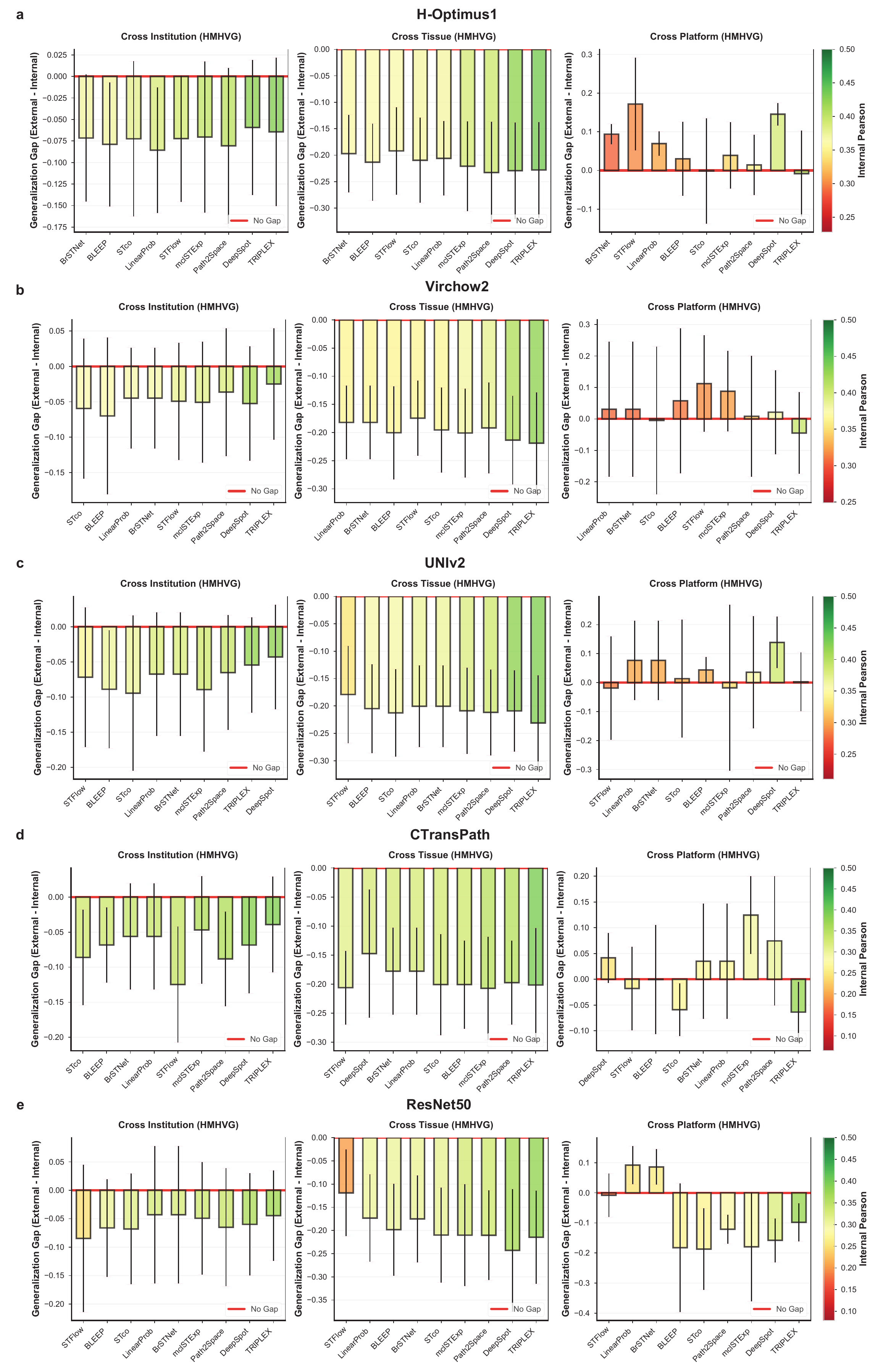}
\refstepcounter{extendeddatafigure}
\caption*{\textbf{Extended Data Fig.~\theextendeddatafigure: Impact of pathology foundation model choice on the generalization gap of ST prediction models}. \textbf{(a-e)} Bar charts quantifying the generalization gap under three generalization scenarios (Cross-Institution, Cross-Tissue, and Cross-Platform), evaluated separately for five patch encoders: H-Optimus1, Virchow2, UNIv2, CTransPath, and ResNet50. Only models compatible with interchangeable patch encoders are included. Models are ordered by increasing Pearson correlation coefficient on STP-BENCH-INTERNAL, also represented by the color code. The red line indicates no generalization gap. Error bars represent standard deviation across all dataset pairs within each scenario.}
\label{fig:generalization_gap}
\end{figure*}

\begin{figure*}
\centering
\includegraphics[width=0.97\textwidth]{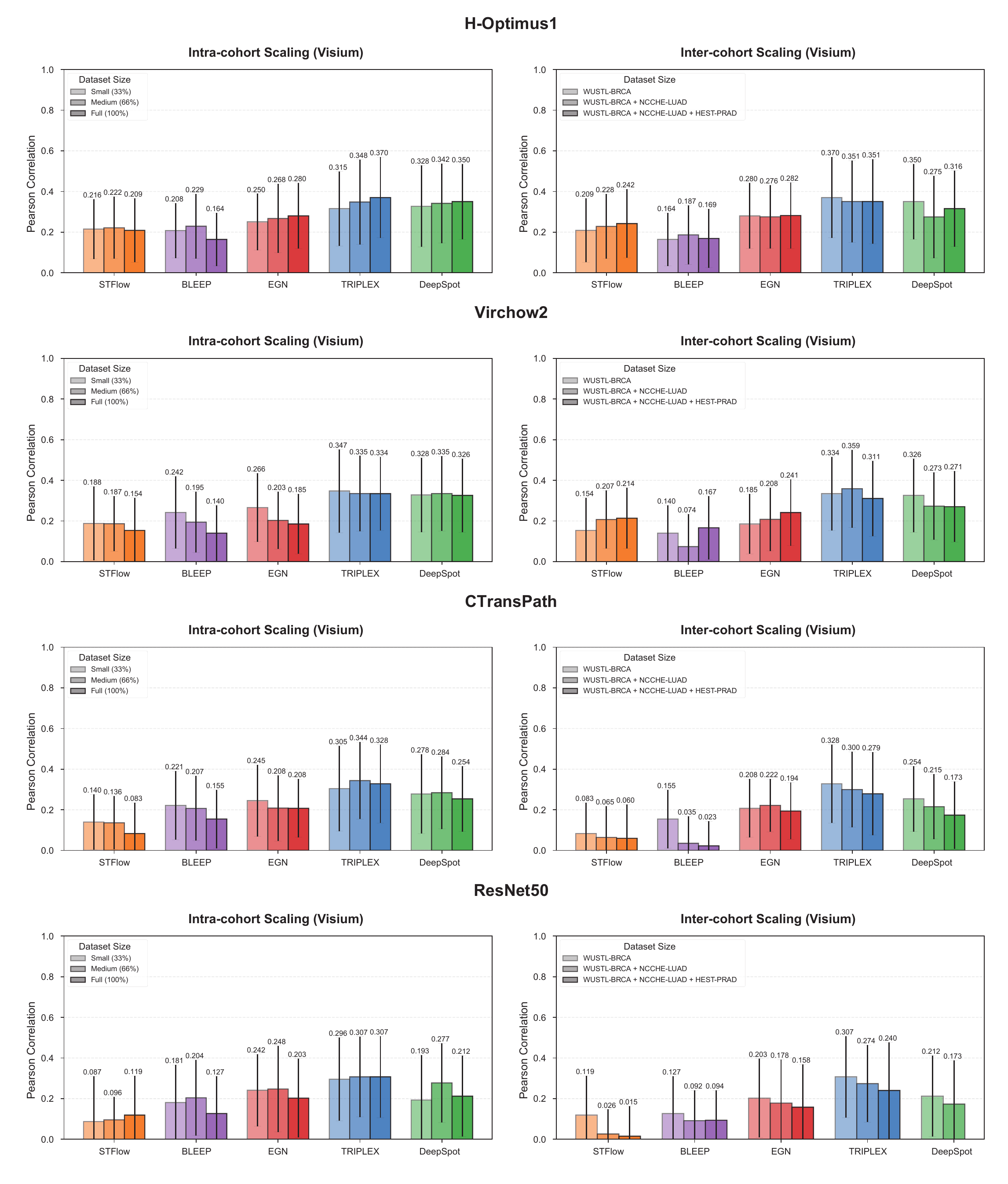}
\refstepcounter{extendeddatafigure}
\caption*{\textbf{Extended Data Fig.~\theextendeddatafigure: Effect of Pathology Foundation Model Choice on Data Scaling}. Scaling performance of selected spatial transcriptomics prediction models evaluated across different pathology foundation models (PFMs) used as patch encoders. The figure shows intra-cohort scaling results on WUSTL-BRCA-Visium using progressively larger training subsets and inter-cohort scaling results with progressively expanded training data composition. Model performance is reported as Pearson correlation, with error bars indicating the standard deviation across samples. These analyses assess whether the observed scaling trends are consistent across different PFM choices.}
\label{fig:scale_pfm}
\end{figure*}

\clearpage
\newcounter{extendeddatatable} 
\renewcommand{\theextendeddatatable}{\arabic{extendeddatatable}} 

\captionsetup{width=\textwidth}
\newcolumntype{L}{>{\raggedright\arraybackslash}X}
\newcolumntype{C}[1]{>{\centering\arraybackslash}p{#1}}
\begin{table*}[t]
\centering
\renewcommand{\arraystretch}{1.15}
\refstepcounter{extendeddatatable}
\setlength{\tabcolsep}{6pt}
\small
\begin{tabularx}{\textwidth}{L C{1.2cm} C{1.2cm} C{1.2cm} C{1.0cm} C{1.0cm} C{1.5cm}}
\toprule
\textbf{Dataset} &
\textbf{Source} &
\textbf{Cancer type} &
\textbf{Platform} &
\textbf{Slides} &
\textbf{Patients} &
\textbf{Total spots} \\
\midrule
\multicolumn{7}{l}{\textit{\oursint}} \\
\addlinespace[2pt]
HEST-PRAD-Visium  & HEST-1K & PRAD & Visium & 23 & 2 & 62,710  \\
WUSTL-BRCA-Visium       & WUSTL   & IDC  & Visium & 47 & 21 & 157,138 \\
NCCHE-LUAD-Xenium     & NCCHE   & LUAD & Xenium & 16 & 16 & 34,519  \\
NCCHE-LUAD-Visium     & NCCHE   & LUAD & Visium & 43 & 30 & 96,954  \\
HEST-CCRCC-Visium & HEST-1K & RCC  & Visium & 24 & 24 & 74,220  \\
SNU-GBM-Visium          & SNU     & GBM  & Visium & 30 & 17 & 117,036 \\
WUSTL-PDAC-Visium       & WUSTL   & PAAD & Visium & 19 & 14 & 66,447  \\
\addlinespace[2pt]
\midrule
\multicolumn{7}{l}{\textit{\oursext}} \\
\addlinespace[2pt]
MGB-PRAD-Visium       & MGB     & PRAD & Visium & 8  & 8  & 31,438  \\
HEST-IDC-Xenium   & HEST-1K & IDC  & Xenium & 4  & 4  & 35,536  \\
Massey-BRCA-Visium      & Massey  & IDC  & Visium & 43 & 22 & 55,227  \\
HEST-LUNG-Xenium  & HEST-1K & LUAD & Xenium & 2  & 2  & 5,206   \\
WUSTL-RCC-Visium        & WUSTL   & RCC  & Visium & 10 & 10 & 33,090  \\
HEST-GBM-Visium         & HEST-1K & GBM  & Visium & 3  & 2  & 17,763  \\
HEST-PAAD-Xenium  & HEST-1K & PAAD & Xenium & 3  & 3  & 7,571   \\
\bottomrule
\end{tabularx}
\normalsize
\caption*{\textbf{Extended Data Table~\theextendeddatatable: \ours{} benchmark dataset composition.} Summary of training (\oursint) and independent test (\oursext) cohorts, detailing data source, cancer type, ST platform, number of tissue slides and patients, and total capture spots per dataset. The internal cohort spans seven datasets across six cancer types and two ST platforms (Visium and Xenium). The external cohort comprises datasets and samples that are non-overlapping with the training data to enable unbiased generalizability assessment.}
\label{tab:dataset_composition}
\end{table*}

\begin{table}[p]
\centering
\resizebox{\textwidth}{!}{

\refstepcounter{extendeddatatable} 
\label{tab:st_gene_prediction_models}
\begin{tabular}{
p{2.0cm} p{2.6cm} p{2.2cm} p{3.2cm} c
p{3.6cm} p{3.2cm} p{3.2cm} p{2.2cm}
p{3.8cm} p{2.8cm} p{2.6cm} p{4.2cm}
}
\toprule
Model &
Input type (histology structure) &
Learning approach &
Core model architecture &
Foundation model available &
Baseline model &
Metric &
Dataset &
Data platform &
Gene selection (number) &
Sample num (Patient num) &
Patch num &
Disease status \\
\midrule

ST-Net &
Target &
Regression &
DenseNet-121 &
X &
-- &
PCC, RMSE &
ST-Net dataset &
ST &
HEG (250) &
68 (23) &
30,612 &
Breast cancer \\

HisToGene &
Target / Global &
Regression &
Transformer &
O &
ST-Net &
PCC &
Her2+ / cSCC &
ST / ST &
HVG (785) / HVG (134) &
32 (7) / 12 (4) &
13,594 / 8,671 &
Breast / Skin cancer \\

EGN &
Target / Retrieval &
Regression &
Transformer &
O &
ST-Net, NSL, ViT, MPViT, CycleMLP, Retro, ViTExp &
MSE, MAE, PCC@F/S/M &
ST-Net dataset &
ST &
HEG (250) &
68 (23) &
30,612 &
Breast cancer \\

BrSTNet &
Target &
Regression &
CNN, Transformer &
O &
ResNet, Inception, EfficientNet, ViT &
MAE, RMSE, PCC &
ST-Net dataset &
ST &
HEG (250) &
68 (23) &
30,612 &
Breast cancer \\

CFANet &
Target / Global &
Regression &
CNN, Attention &
O &
ST-Net, NSL, ViT, CycleMLP, MPViT, EGN, EGGN &
MSE, MAE, PCC@F/S/M &
ST-Net dataset / 10x Genomics &
ST / Visium &
HEG (250) &
68 (23) / 6 &
30,612 / 32,032 &
Breast cancer \\

EGGN &
Target / Retrieval &
Regression &
GNN &
O &
ST-Net, NSL, ViT, CycleMLP, MPViT, Retro, ViTExp, EGN &
MSE, MAE, PCC@F/S/M &
ST-Net dataset &
ST &
HEG (250) &
68 (23) &
30,612 &
Breast cancer \\

SEPAL &
Target / Local &
Regression &
GNN &
O &
ST-Net, EGN, EGGN, HisToGene &
MAE, MSE, PCC-Gene/Patch, $R^2$-Gene/Patch &
ST-Net / 10x BC &
ST / Visium &
High Moran's I genes (256) &
68 (23) / 1 (2) &
30,612 / 7,785 &
Breast cancer \\

BLEEP &
Target &
Bi-modal \& Retrieval &
CLIP &
O &
ST-Net, HisToGene &
PCC &
Human liver &
Visium &
Marker (8), HEG (50), HVG (50) &
4 (1) &
9,269 &
Human liver tissue \\

M2ORT &
Target (multiple) &
Regression &
Transformer &
X &
DeepSpaCE, ST-Net, HisToGene, Hist2ST &
PCC, RMSE &
ST-Net / Her2+ / cSCC &
ST / ST / ST &
HEG (250) &
68 (23) / 36 (8) / 12 (4) &
30,612 / 13,594 / 8,671 &
Breast / Skin cancer \\

SGN &
Target / Text &
Regression &
CNN, GNN, Transformer &
O &
ST-Net, NSL, EGN, HSANet, CFNet, EGGN &
MSE, MAE, PCC@F/S/M &
ST-Net dataset &
ST &
HEG (250) &
68 (23) &
30,612 &
Breast cancer \\

TRIPLEX &
Target / Local / Global &
Regression &
Transformer &
O &
ST-Net, EGN, BLEEP, HisToGene, Hist2ST &
MSE, MAE, PCC (M/H) &
ST-Net / Her2+ / cSCC / 10x &
ST / ST / ST / Visium &
HEG (250) &
68 (23) / 36 (8) / 12 (4) / 3 (3) &
68,050 / 13,620 / 23,205 &
Breast / Skin cancer \\

mclSTExp &
Target / Coord &
Bi-modal \& Retrieval &
CLIP &
O &
ST-Net, HisToGene, His2ST, THItoGene, BLEEP &
PCC (ACG/HEG), MSE, MAE &
Her2+ / cSCC / Alex+10x &
ST / ST / Visium &
ACG, HEG (top 50) &
36 (8) / 12 (4) / 9 &
13,594 / 8,671 &
Breast / Skin cancer \\

STco &
Target / Coord &
Bi-modal \& Retrieval &
CLIP &
O &
ST-Net, HisToGene, His2ST, THItoGene, BLEEP &
PCC &
Her2+ / cSCC &
ST / ST &
ACG, HEG (top 50) &
32 (7) / 12 (4) &
-- &
Breast / Skin cancer \\

HistoSPACE &
Target &
Regression &
Autoencoder &
X &
ST-Net, Hist2ST &
PCC &
ST-Net / Her2+ &
ST / ST &
HEG (250), GNAS, HLA-B &
68 (23) / 36 (8) &
-- &
Breast cancer \\

Path2Space &
Target &
Regression &
Foundation model + MLP &
O &
Hist2ST, HisToGene &
PCC &
Her2+ &
ST &
Top 50 predicted genes &
36 (8) &
-- &
Breast cancer \\

M2OST &
Target (multiple) &
Regression &
Transformer &
X &
ST-Net, DeepSpaCE, HisToGene, Hist2ST, BLEEP, HIPT/iStar &
PCC, RMSE &
ST-Net / Her2+ / cSCC &
ST / ST / ST &
HEG (250) &
68 (23) / 36 (8) / 12 (4) &
30,612 / 13,594 / 8,671 &
Breast / Skin cancer \\

DeepSpot &
Target / Local &
Regression &
Foundation model + MLP &
O &
BLEEP, ST-Net, HisToGene, Hist2ST &
PCC &
Tumor Profiler, HEST-1K &
Visium &
HVG (5000) &
18 (7) / 24 (24) / 8 (4) / 6 (6) &
-- &
Melanoma, Kidney, Colon \\

Stem &
Local &
Generative &
Diffusion &
O &
Hist2ST, BLEEP, TRIPLEX &
PCC, MAE, MSE, RVD &
Kidney, Her2+, Prostate, Mouse &
Visium / ST / Visium / Visium &
HMHVG(200) / HMHVG(300), DEG(296), DEG(1000) / HMHVG(200) / HMHVG(200) &
23 (22) / 36 (8) / 23 (2) / 14 (4) &
-- &
Kidney / Breast / Prostate / Mouse \\

STFlow &
Global &
Generative &
Flow matching &
O &
Ciga, UNI, GigaPath, BLEEP, TRIPLEX &
PCC &
HEST-1K, STImage-1K4M &
Visium, Xenium &
HVG (50) &
74 (48) / 146 &
-- &
Multiple cancers \\

OmiCLIP &
Target &
Bi-modal \& Retrieval &
CLIP &
-- &
GigaPath, UNI, Hist2ST, BLEEP &
MSE, PCC &
ST-bank &
Visium &
HEG (300) &
1007 &
2.2M &
Normal / Cancer \\

STPath &
Target &
Generative &
Spatial transformer &
-- &
UNI, GigaPath, BLEEP, TRIPLEX &
PCC, AMI &
HEST-1K, STImage-1K4M &
Visium, Xenium, ST &
Test: HVG (50), Valid: HVG (200) &
983 &
-- &
Multiple cancers \\

\bottomrule
\end{tabular}
}
\caption*{\textbf{Extended Data Table~\theextendeddatatable:
Comparison of ST prediction models.}
FM indicates whether a foundation model is used.
HEG: highly expressed genes; HVG: highly variable genes;
HMHVG: high-mean highly-variable genes; ACG: all considered genes.
}

\end{table}

\captionsetup{width=\textwidth}

\newcolumntype{L}{>{\raggedright\arraybackslash}X}
\newcolumntype{C}[1]{>{\centering\arraybackslash}p{#1}}

\begin{table*}[t]
\centering
\renewcommand{\arraystretch}{1.15}

\refstepcounter{extendeddatatable} 
\setlength{\tabcolsep}{4pt} 
\small 

\begin{tabularx}{\textwidth}{p{2.1cm} C{1.5cm} L C{2.1cm} C{1.2cm} L C{1.3cm} C{1.2cm}}
\toprule
\textbf{Model} &
\textbf{Pretrain size} &
\textbf{Pretrain dataset detail} &
\textbf{Magnification} &
\textbf{SSL} &
\textbf{Model architecture} &
\textbf{Model size} &
\textbf{Year} \\
\midrule
H-Optimus1 & 1M+   & 50 organs, 800,000 patients & 20x & DINOv2 & ViT-g/14 & 1.1B & 2025 \\
Virchow2   & 3.1M  & 150+ tissues, 225,401 patients, H\&E + IHC & 5x, 10x, 20x, 40x & DINOv2 & ViT-H/14 & 632M & 2024 \\
UNIv2      & 350k+ & 20+ organs, H\&E + IHC & 20x & DINOv2 & Modified ViT-\allowbreak H/14 & 681M & 2024 \\
CTransPath & 30k   & 32 cancer subtypes, 10,953 patients & 10x & SRCL & CNN+\allowbreak Swin-\allowbreak Transformer & 28M & 2021 \\
\bottomrule
\end{tabularx}

\normalsize
\caption*{\textbf{Extended Data Table~\theextendeddatatable: Pathology foundation model (PFM) description.} Information about each PFM (parameters, pretraining data size, release year, etc.).}
\label{tab:pretrain_models}
\end{table*}

\clearpage

\end{spacing}
\newpage
\begin{nolinenumbers}
\Heading{References}
\bibliographystyle{nature}
\bibliography{main}
\end{nolinenumbers}




\end{document}